\documentclass[12pt]{article}
\usepackage[margin=0.8in,top=1in]{geometry}
\usepackage[T1]{fontenc}
\usepackage[utf8]{inputenc}
\usepackage{fancyhdr}
\usepackage{lastpage}
\usepackage{graphicx, wrapfig, subcaption, setspace, booktabs}
\usepackage[font=small, labelfont=bf]{caption}
\usepackage{natbib}
\usepackage{hyperref}
\usepackage{fourier}
\usepackage[protrusion=true, expansion=true]{microtype}
\usepackage[bottom,hang,flushmargin]{footmisc}
\usepackage{url, lipsum}
\usepackage{subcaption}
\usepackage{amsmath,amssymb}
\usepackage{graphics}
\usepackage{graphicx}
\usepackage{algorithm}
\usepackage{algorithmicx}
\usepackage{caption}
\usepackage{xcolor}
\usepackage{mathtools}
\usepackage{comment}
\usepackage{enumitem}
\usepackage{wrapfig}

\newcommand{\HRule}[1]{\rule{\linewidth}{#1}}
\newcommand\blfootnote[1]{%
  \begingroup
  \renewcommand\thefootnote{}\footnote{#1}%
  \addtocounter{footnote}{-1}%
  \endgroup
}

\newcommand{\customfootnotetext}[2]{{
  \renewcommand{\thefootnote}{#1}
  \footnotetext[0]{#2}}}

\begin{document}\thispagestyle{empty}


\noindent
\begin{minipage}[t]{0.5\textwidth}
Department of Computer Science\\
Rochester Institute of Technology\\
\texttt{https://www.rit.edu/computing/}
\end{minipage}
\hfill
\begin{minipage}[t]{0.5\textwidth}
\begin{flushright}
1 Lomb Memorial Dr\\
Rochester, NY 14623\\
\end{flushright}
\end{minipage}
\HRule{1pt}

\vspace{4.5cm}

\begin{center}
August 12, 2026\\ 
\noindent\rule{14cm}{2.5pt} \\
NAC TR 24237-1\\ 
\vspace{0.35cm}
\noindent
{\fontsize{18}{12}\selectfont \textbf{A Practical Guide to Tuning Spiking Neuronal Dynamics for Computational Neuroscience and NeuroAI Research}} \\ 
{\fontsize{10}{12}\selectfont Version 1} \\
\vspace{0.35cm}
\textbf{William Gebhardt\textsuperscript{$*$}, Alexander G. Ororbia II\textsuperscript{$*$}, Nathan McDonald, \\Clare Thiem, Jack Lombardi, Cory Merkel} \\ 
The Neural Adaptive Computing Laboratory\\
Air Force Research Laboratory
\noindent\rule{14cm}{2.5pt} \\
\end{center}

\newpage\thispagestyle{empty}
\begin{center}
\noindent\rule{17.5cm}{1.5pt} \\
\vspace{0.75cm}
\noindent
{\fontsize{16}{12}\selectfont \textbf{A Practical Guide to Tuning Spiking Neuronal Dynamics for Computational Neuroscience and NeuroAI Research}} \\
{\fontsize{10}{12}\selectfont Version 1} \\
\vspace{0.25cm}
\noindent\rule{17.5cm}{1.5pt} \\
\textbf{William Gebhardt\textsuperscript{$*$}, Alexander G. Ororbia II\textsuperscript{$*$}, Nathan McDonald, \\Clare Theim, Jack Lombardi, Cory Merkel} \\ 
The Neural Adaptive Computing Laboratory\\
Department of Computer Science, Rochester Institute of Technology
\end{center}

\customfootnotetext{$*$}{Authors contributed equally.}

\tableofcontents\blfootnote{
The views and conclusions contained herein are those of the authors and should not be interpreted as necessarily representing the official policies or endorsements, whether expressed or implied, of the Air Force Research Laboratory (AFRL) or the U.S. Government. This material has been cleared for public release, unlimited distribution. (PA Case \#: AFRL-2026-3409).}

\newpage 

\section{Introduction}
\label{sec:intro}
There are countless papers, efforts, and methods that utilize spiking neural networks (SNNs) to solve various modeling problems \cite{gerstner1998spiking,maass1999computing,izhikevich2006polychronization,yamazaki2022spiking,ororbia2023spiking,gebhardt2024time,n2024predictive,hu2024advancing,gygax2025elucidating}. Each of these efforts generally use their own specific formulation of the SNN topology and its learning process in order to achieve success with respect to specific research goals. Nevertheless, many of these SNN-centric efforts share, at least in part, one element in common: their use of foundational components and concepts, such as the basic neuronal cell building block or a means of data encoding. 
As these basic ingredients are found across computational neuroscience and brain-inspired computing \cite{parhi2020brain,yamazaki2022spiking} efforts, in the best of cases, the knowledge of the hyper-parameters used, how to tune them, as well as the overall dynamical changes caused by alterations made to these hyper-parameter values has to regularly be gleaned in small doses from dozens of different articles. This makes it difficult for newcomer researchers and students, coming from a variety of neuroscientific and neuromorphic backgrounds, to readily and accessibly build and simulate their own effective SNN models.  
In this work, we move to provide an overview of these common components as well as create a useful perspective and guide on configuring and tuning common parameters of key building blocks that constitute SNN models. 

In this article, we will study different key methods commonly used to encode real-valued data values into a format that is usable by SNNs, including both probabilistic and periodic methods, i.e., Bernoulli trials, Poisson processes, and phasors. We will further break down the formulation of two key neuronal building block models: the leaky integrate-and-fire (LIF) neuron and the resonate-and-fire (RAF) neuron. Furthermore, we will examine how certain hyper-parameters affect LIF and RAF dynamics and then, finally, we will examine a small excitatory-inhibitory assembly of LIF units -- an important motif found within larger SNN models. 
Such an assembly will allow us to study a basic SNN that embodies many of the key the dynamics that a modeler might come to desire and expect.

\noindent 
\textbf{Motivation: }
We write this article to serve as a teaching tool rather than as an effort to fill a particular technical gap in brain-inspired computing or computational neuroscience. In other words, our contribution is the development of a much-needed practical guide\footnote{Code to run many of the key simulations, using the computational neuroscience and NeuroAI Python library NGC-Learn, has been made publicly available in the public Github repository: \texttt{https://github.com/NACLab/SnnDynamicsPracticalGuide}} for newcomer researchers and practitioners who need to build well-working SNN systems themselves, either for use in studies in computational neuroscience or application contexts in neuroscience-motivated artificial intelligence (NeuroAI) \cite{zador2023catalyzing}, neuroscience-motivate machine learning (NeuroML) \cite{ororbia2024review}, and brain-inspired computing \cite{kendall2020building}. 
Our motivation in writing this article is thus twofold: 
\begin{itemize}[noitemsep,nolistsep,topsep=0pt]
    \item \textbf{1)} We seek to provide a central piece that lays the groundwork for tuning certain basic building-block components in SNNs, and 
    \item \textbf{2)} We develop a guide enable those who are new to SNNs in building their first biophysical models without having to dissect large bodies of literature in order to gather all of the requisite practical information.
\end{itemize}
Note that this work will not be covering how to train an SNN, e.g., via Hebbian learning or spike-timing-dependent plasticity \cite{bi1998synaptic,markram2012spike}, or dealing with the dynamics of an evolving system \cite{ororbia2023brain}; instead, we will focus on setting up the dynamics for a randomly initialized network. Our rationale for ensuring that a non-evolved, ``un-trained'' SNN is properly working is that, if the initial system is not functioning in an expected manner (it exhibits ``unhealthy'' behavior) then it would be unreasonable to expect that the same system evolving over time would function or operate correctly (and furthermore, the learning process could possibly result in further degenerate behavior).  

\section{Biological Foundations and Dynamics}
\label{sec:method}
In computational neuroscience \cite{gerstner2014neuronal} and brain-inspired computing \cite{strukov2019building,ororbia2023brain,li2024brain}, spiking neural networks (SNNs), often referred to as the third generation of artificial neural networks (ANNs) \cite{maass1997networks}, are computational models that attempt to mimic natural neuronal networks and, in the context of NeuroAI \cite{marblestone2016toward,zador2023catalyzing}, are regarded as a form of neuro-mimetic information processing \cite{friston2010free,ororbia2024review}. Beyond modeling some degree of neuroanatomy \cite{frackowiak1994functional,fix2002neuroanatomy} in terms of neuronal (membrane potential) and synaptic state (the synaptic cleft), 
SNNs operate under the notion that neuronal units within the SNN do not transmit information at each propagation cycle and, instead, only fire when their membrane potential (a characteristic of a biological neuronal cell related to its membrane's electric charge) crosses a specific value threshold. Once a neuron fires, a pulse signal -- an action potential -- is communicated / transmitted to other connected neurons, which either decreases or increases their own respective membrane potentials \cite{o2000computational,gerstner2014neuronal}. Ultimately, the above means that SNN models inherently center around the timing of the discrete events -- the spikes -- emitted by the neuronal units that constitute it; in other words, 
the key signaling mechanism through which biological neurons communicate over long distances is the spike train\footnote{Such a train, in continuous time, is generally represented as a series of timestamps where the pulses occur. In discrete time simulation, this is instead computationally represented as an array of zeros and ones, where each element is a time step and the spikes occur at nonzero elements. Biologically, since action potentials do not change as they propagate along axons, their particular shapes and amplitudes carry no information (isolated spikes for a specific neuron look much alike). This ultimately means that it is the number and the timing of the spikes within a spike train that effectively matter.}. 
Furthermore, spike trains, 
which carry high information coding capacity \cite{taherkhani2020review}, facilitate the high-speed processing that humans and animals are capable of; for instance, in the human visual system, image recognition can be performed in as little as $13$ milliseconds of processing time \cite{potter2014detecting}, 
going through multiple cortical layers (from retina to the temporal lobe). 


\begin{figure}[!t]
    \centering
    \includegraphics[width=0.4\textwidth]{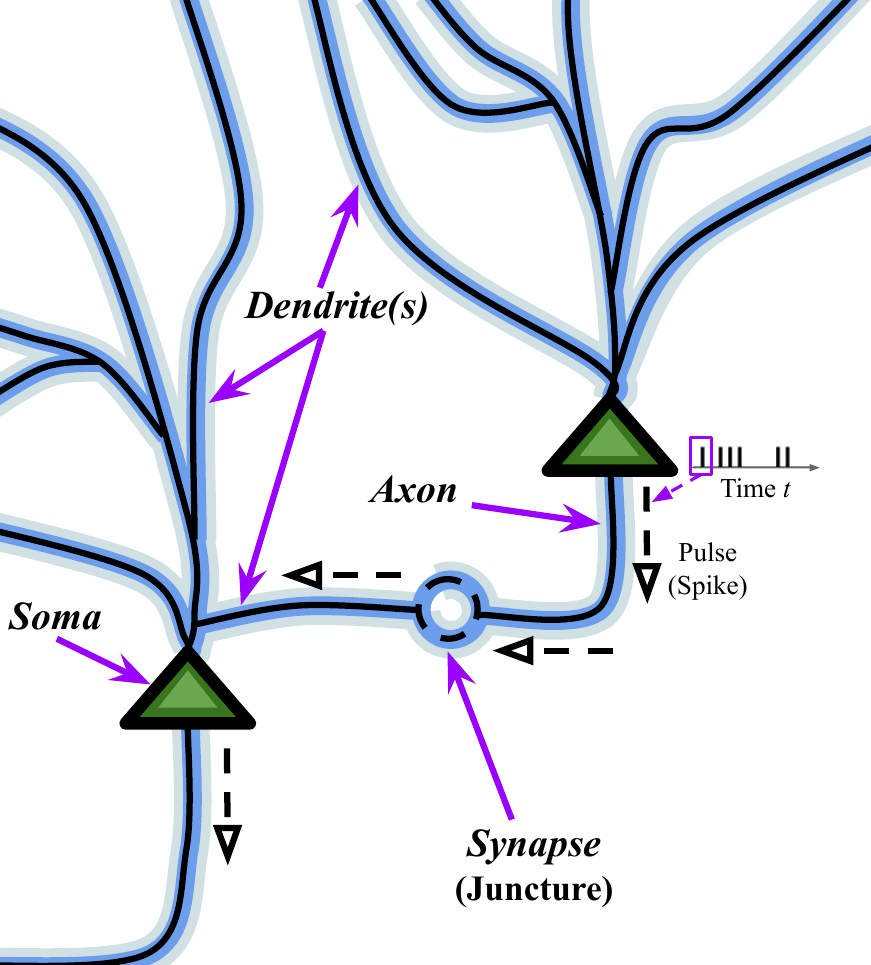}
    \caption{
    Depiction of natural neuronal cells. One neuron (to the right) transmits pulses that feed into a second one (to the left). Labeled are the core 
    morphological neural components: 
    the dendrites (the site of input processing), 
    the soma (the site of nonlinear integration of dendritic-processed values), and 
    the axon (the output transmitter). 
    }
    \label{fig:bio_neurons}
\end{figure}

In the brain, most natural neurons (Figure \ref{fig:bio_neurons}) can be generally divided into three functionally distinctive parts: 
\textbf{1)} the dendrites (the dendritic tree), 
\textbf{2)} the soma, and, 
\textbf{3)} the axon (axonal branches). Dendrites largely play the role of input integration and collect the pulse signals emitted by other neurons and then transmit the result to the soma. The soma itself can be treated as a (central) processing unit that carries out nonlinear processing of what the dendrites provide it, i.e., if the total input that arrives at the soma exceeds some particular threshold, then it decides that an output signal must be produced. This output signal is then taken over by the axon, or the ``output device'', which ultimately delivers this signal to other neurons. Note that the junction (or gap / space) between two neurons is called the synapse -- the sending neuron (before the synaptic gap) is referred to as the pre-synaptic cell whereas the receiving neuron (after the synaptic gap) is referred to as the post-synaptic cell. Note that one neuron (in something such as the vertebrate cortex) can connect up to more than $10,000$ post-synaptic neurons; the neuron's axonal branches can stretch over several centimeters in order to reach other neurons in the brain. 

Many SNN models in computational neuroscience often attempt to emulate some aspects of the above neuro-machinery, morphology, and behavior, typically providing a more detailed mathematical description of the soma's nonlinear integration of values received by the dendrites as well as the emission of action potentials across the axon (with more recent, increasing interest in building more detailed models/machinery of the dendritic tree itself \cite{jones2021might,pagkalos2023introducing}). 
A large plethora of mathematical models exist, ranging from more complex, neuroanatomically-faithful ones such as that proposed by Hodgkin and Huxley \cite{hodgkin1952quantitative} to the computationally-fast-to-simulate yet vastly simplified leaky integrator \cite{stein1965theoretical}. However, while all models are wrong \cite{box1976science}, failing to perfectly capture the complexity of real natural neurons, some prove to be useful, describing particular subsets of behaviors, properties, or characteristics of such cells and their underlying dynamics. As a result, it is common in both the enterprises of computational neuroscience and brain-inspired computing to make use of vastly simple mathematical neuronal models -- and modifications of them -- such as the leaky integrate-and-fire \cite{stein1965theoretical} (LIF) and resonate-and-fire \cite{izhikevich2001resonate} (RAF) neuron. Notably, the LIF and RAF often serve as fast-to-simulate building blocks for more intricate, sophisticated systems; this results in the design of energy-efficient neuro-mimetic models that are useful for applications in NeuroAI and NeuroML research \cite{zador2023catalyzing,ororbia2024review}. 

We next move to cover, briefly, three important background  elements from computational neuroscience -- the concept of burst firing, dynamical systems, and time reliance -- that motivate the mathematical and computational models studied in this guide.

\subsection{Burst Firing}
Burst firing, or bursting, is an activation pattern observed in particular neurons \cite{izhikevich2000neural,gerstner2014neuronal}, particularly where periods of rapid spiking are followed by longer periods of quiescence (the length of the quiescent period is much longer than the neuron's inter-spike interval). With respect to terminology, spikes bursts can be broken down into: 
\textit{1)} a doublet, or two-spike burst, 
\textit{2)} a triplet, or three-spike burst, and 
\textit{3)} a quadruplet, or four-spike burst. %
Biologically, there are two general kinds of bursting neurons: input-driven bursting, where a cell exhibits burst firing behavior if it is driven by a constant (subthreshold) input current, and intrinsic bursting, where a cell produces a bursting pattern rather independently of the input (often due to complex feedback patterns). Often, a neuronal cell can be treated as two coupled sub-systems \cite{izhikevich2000neural} (or evolving variables) to model bursting behavior; a fast sub-system responsible for each individual pulse emitted and a slow sub-system that characterizes the shape and intensity of the spikes before quiescence is triggered. As we will observe later in this article, the resonate-and-fire neuron is one such model that embodies a fast and slow sub-system using two linked differential equations. 

\subsection{Dynamical Systems}
A dynamical system \cite{birkhoff1927dynamical} is essentially a function that describes the temporal or time dependence of a point in space. Generally, the dynamics (how something changes with respect to time) of a dynamical system can be described via ordinary differential equations, further specifying how the system's properties in space and time might be measured. Specifically, at any given time, a dynamical system can be explained as having a state that represents a point within its overall state-space; the state is often a vector containing numbers that describe particular features of the system. The equations that describe the system's motion (dynamics) make up the system's evolutionary behavior or trajectory in state-space, i.e., they describe what future states might follow from the current state. To advance from one state to another, the underlying differential equations that explain the system must be iterated, often via a small time step; this means that integration of the system's underlying differential equations is required. Note that using ordinary differential equations (ODEs) to describe the motion of a system makes it a continuous dynamical system (whereas using difference equations instead makes it a discrete system). 

In the context of this article, we consider a spiking network to be a 
dynamical system \cite{eliasmith2003neural} that is simulated in discrete time (i.e., its ODEs are integrated via schemes such as Euler integration \cite{euler1768OperaOmnia}), given that, even in its most basic format, the neuronal units that compose it are typically described using the language of differential equations. As a result, describing an SNN means that its operation is viewed as happening over or across a period of time, rather than one single instant point. This stands in contrast to most modern-day deep neural networks (DNNs) which technically only ``exist'' within an instant of time and only when data is presented to them; in other words, a DNN does not ``exist'' until data is propagated through it (in a feedforward pass) whereas an SNN does not require data input in order to exist (its units operate continually, even in the absence of input stimuli). Ultimately, this means that an SNN can be viewed, in a senese, as a (dynamical) system in motion, where stimuli (data patterns, representative of an environment) perturb or are injected into its trajectory.  

\subsection{Considering Time-Scales}
Due to the way that SNNs function (motivated further by the fact that, as described above, they can treated as dynamical systems that move through time), they serve as natural candidates for operating  with time series data, processing sensory input as a stream rather better than extrapolating 
from single points. This follows biology in a much closer fashion as sensory input that enters a biological system is not truly a point-wise, static input; it is instead a ``flow'' that is processed over a quantity of time at differing speeds (based on the system processing the information). Knowing how to adapt models to different time-scales is something that is essential for modeling biological systems effectively.

Many equations and dynamics of these systems, which are integrated over (simulated) time utilizing various methods \cite{euler1768OperaOmnia,runge1895,kutta1901},  have the concept of a \textit{time-step}, or change in time, often denoted as $\Delta t$. Many biological systems, such as SNNs, are often contextualized as functioning on a time-scale of milliseconds (ms); however, some models work in terms of other fractions of seconds, e.g., nanoseconds, instead (sometimes this is motivated by the hardware platform that is to instantiate the SNN). Knowing what the units are that the models and their constituent components work under is essential to using and understanding how to convert between the two (particularly when certain neuronal models are formulated in terms of one time-scale but the goal is to craft an SNN at a different time-scale).

It is important to note that, since all of these systems are grounded in mathematical derivations and mechanics, it will always be possible to convert (a configuration of model coefficients, parameters, and/or hyper-parameters) from one time-scale to another, and often identically functioning models can be made with vastly different time-scales. For instance, if coefficient values are chosen and tuned for a time-scale that is on the order of $50$ ms, there often exists another set of values that would work out identically when tuned to the order of $5$ ms. 
Solving for these other values (or configurations of them) is not often needed; nevertheless, when reading work from many different authors and/or research teams, it is likely that these different studies employ different time-scales \cite{zeraati2024neural}, which can make clear and direct experimental comparison and reproduction difficult \cite{manninen2018challenges} (this is not just problem confined to computational neuroscience or NeuroAI, as it is a problem for scientific, empirical work in general \cite{trevors2010perspective}). As a result, at least some translation or conversion will be required for the experimentalist or practitioner.

\section{Input Encoding}
The first step in designing many SNN models is to consider the scheme for encoding input data. In many cases, the input data is a snapshot of an environment, fixed in time. An SNN, however, moves through time as all of its internal dynamics are driven by stimuli. Due to this natural temporal flow, it becomes necessary to transform or encode these fixed values into a range of states that cover the simulation time; notably, this is useful for emulating target neuromorphic instantiations of SNNs, such as those that will process the event streams emitted by dynamic vision sensors (DVS) \cite{bi2018spike, prophesee2023dvs}. 
Furthermore, the input stream will need to converted to a discrete, event-based format, since the values that will be propagated through the SNN are to be binary signals (representing either ``on'' or ``off''). 
As a result, if the inputs provided to the model exist as real values, then an additional transformation at each time-step -- specifically one that converts each real value into a binary signal -- will be required. 
In the field, there are many methods for accomplishing this transformation/conversion, often dependent on the input data's dynamics as well as the overall meaning of the sensory to be modeled; in the following sections, we will discuss and study several of the most common mechanisms for converting real-valued patterns to spike trains. Note that, in the event that the inputs are sequences of real-valued patterns (e.g., frames in a video sequence), then the processing will need to further convert each pattern at each step to a stream of discrete events or take into account the sequence's ordered nature to encode spike pulses.

\subsection{Intensity Encoders}

Whether working in the physical world or in simulation, the use of sensors that measure the intensity of different aspects of an environment is quite common. These sensors are often chosen based on their ability to highlight the ``important'' parts of an image through the magnitude of their output. A common example of this is through the use of cameras and their pixel outputs. In this instance, a white pixel (a value of $255$ or normalized value of $1$) is often referred to as ``high intensity'' whereas a black pixel as ``low intensity'' (a value of $0$). 
This not only applies to cameras but to many other types of sensors as well, such as LIDAR (laser imaging, detection, and ranging) \cite{taylor2019introduction}, which also model their output through intensity. Encoding intensity can be done through a multitude of techniques each with their own strengths and limitations, in this section two such methods will be addressed.

\subsubsection{Bernoulli Encoding}
\label{section:input:normalized:raw}

Treating the intensity as a normalized real value (such a probability to sample from) is the most straight-forward way to encode it into a spike train; this is done through the use of Bernoulli trials. 
At its core, a Bernoulli trial centers around the idea of flipping a weighted coin and checking to see if it comes up as heads. If we treat each time step as its own independent flip of this weighted coin, we obtain a sequence of Bernoulli trials -- the resultant sequence adheres to a Binomial distribution. Formally, this distribution can be written in terms of the probability of exactly $k$ successes out of $n$ trials as:
\begin{equation}
    P(x) = \binom{n}{k}x^k(1-x)^{n-k}.
\end{equation}
Since the above equation is expressed over $n$ trials (or time steps) in the context of our model, this means that, in order to use it to produce a spike train, we will need to produce (pre-compute) the full spike train prior to feeding it (iteratively) in to the SNN model. 
This, however, can prove to be problematic since, normally, the goal is to work with an input encoder that can generate an encoding in sync with the incoming data stream; this is often referred to as ``encoding online'' (or real-time encoding). In order to circumvent this particular limitation, we can instead set the number of successes ($k$) and the trials ($n$) in the Binomial distribution to one. As a result, using these values, the equation for the probability mass function then collapses to:
\begin{equation}
    P(x) = \binom{1}{1}x^1(1-x)^0 = x.
\end{equation}
Thus, we can see that constructing an online binomial distribution (or a Bernoulli trial) for encoding normalized input data collapses to a single weighted coin flip conditioned on the normalized real value input. This is results in a fast sampling process that will produce spike trains on the fly, i.e., we sample $P(x) = x$ each time we want to obtain one time-step slice of the event stream to feed into the input/sensory layer of an SNN.

\subsubsection{Poisson Process Encoding}
\label{section:input:normalized:scaled}

While the above Bernoulli-encoding approach for producing a spike train (from normalized real valued inputs) will result in a valid spike train, there are a few issues that will appear once employing it. One of the more common problems found is that, if the normalized input $x$ is one, or almost one (i.e., $x=1$), the Bernoulli encoder will emit a spike every single time-step. This is, however, not an ideal scenario when working with input streams to be fed into SNN structures. 
Specifically, SNNs rely on the temporal dynamics of the encoded spike trains and, if spikes are emitted at every time step, then these temporal dynamics will be lost or overshadowed. In effect, the SNN's internal neuronal units will be ``over-driven'' by the constant flood of sensory input signals; this would not reflect the real-world sparsity of a stream of values produced by, for example, a low-power DVS. 

To address this, we turn to another observation in biology with respect to how receptors within the body react to high degrees of stimulus. One useful assumption made is that cells emit spike patterns compatible with Poisson processes \cite{amit1997cortex}. Under this assumption, if the goal is to match behavior of realistic neuronal firing rates, adhering generally to a 
Poisson process, then a modification to the above Bernoulli sampling scheme is required.

\begin{figure}
    \centering
    \includegraphics[width=0.7\linewidth]{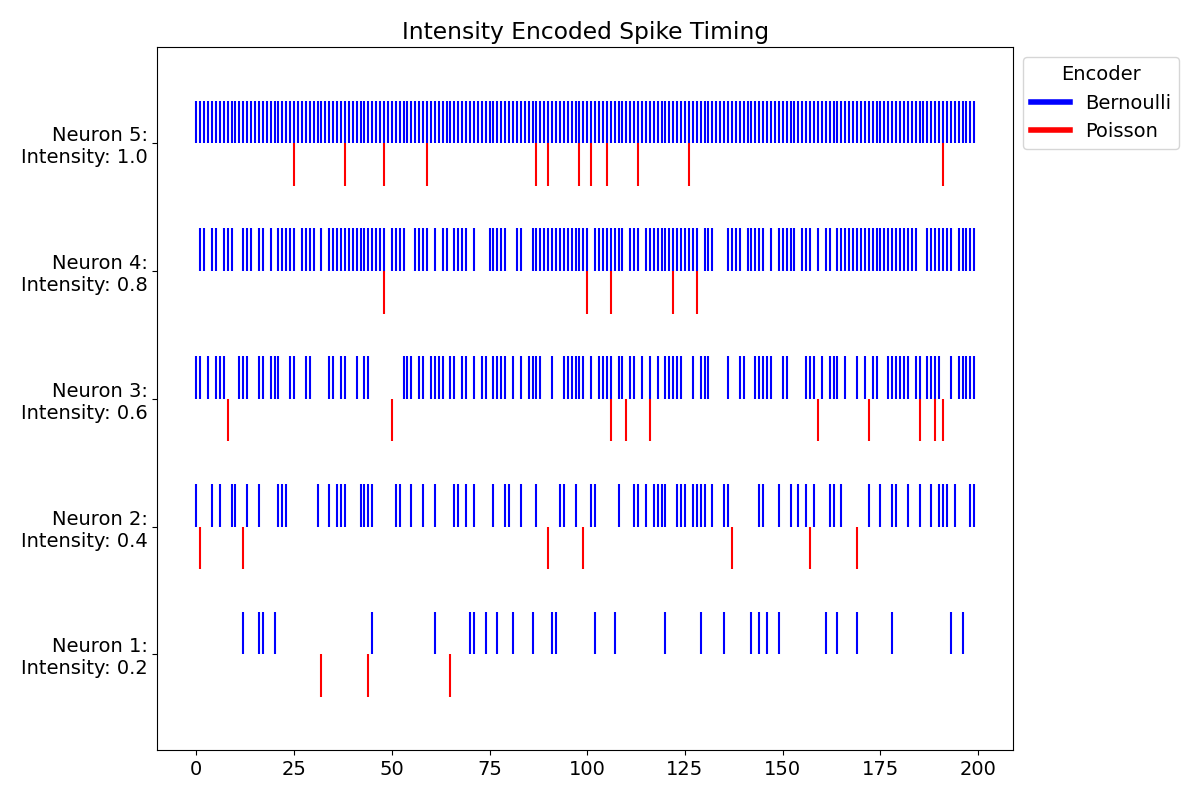}
    \caption{Here the two different encoding schemes (Bernoulli \ref{section:input:normalized:raw} and Poisson process \ref{section:input:normalized:scaled}) can be seen side by side when given the same inputs. For this graph the expected firing rate for the Poisson process is $f_e = 63.75$ \textit{events per second} and a $\Delta t = 1\textit{ms}$. From the figure it can be seen that the Bernoulli encoding produces substantially more stimulus but since so much is produced the fine grain differences between values is lost.}
    \label{fig:input:normalized:comparison}
\end{figure}

To start, let the expected frequency of events for a maximally active encoder be $f_e$. Then, let the normalized real valued input $x$ be a scale factor applied to the expected frequency. Applying this fact, we obtain the following: 
\begin{equation}
    \hat{f_e} = x f_e
\end{equation}
where $\hat{f_e}$ is the scaled expected frequency of the encoder. To finally use this value, the input units then have to be adjusted to align with the sampling rate. Traditionally, $f_e$ will be in terms of \textit{events per second} and thus needs to be converted to \textit{events per timestep} in order to calculate the probability of an event happening at any given time-step. 
This can be done through the following transformation: 
\begin{equation}
    f_s = \hat{f_e} * \dfrac{\Delta t}{1000}
\end{equation}
where $\Delta t$ is in terms of \textit{milliseconds per timestep}. Once $f_s$ is calculated, it can then be used as the probability in a Bernoulli trial in order to produce an encoding online. Starting from the raw values, the explicit equation for the probability of a spike happening at any given timestep given $\Delta t$ and $f_e$ is then: 
\begin{equation}
    P(x; \Delta t, f_e) = x * f_e * \dfrac{\Delta t}{1000}.
\end{equation}
When the above equation is sampled over an entire period, the number of events that happen will follow a Poisson distribution. 
In Figure \ref{fig:input:normalized:comparison}, we visually compare the outputs of a Bernoulli encoder with a Poisson process encoding scheme.

\subsubsection{Limitations of Intensity Encoding}
\label{sec:intensity_limitations}

While the above two methods facilitate a rather easy encoding of intensity values into a spike train, they are not without limitations. The largest of which is that the inter-spike timing (the time between spikes) produced by these approaches exhibits a high variance. This is due to the probabilistic sampling that is used to produce the spike trains themselves -- while this is not necessarily biologically implausible and does not generally cause simulation issues, during model fitting/training (adaptation of synaptic efficacies) it can result in a loss of information. If the input's intensity is low or changing, it can become difficult for these methods to accurately capture a useful representation of this. This stochasticity in these encoders' outputs also limits the efficacy of certain neuronal models as some require or expect more periodic signals (e.g., SNNs composed of units that adhere to oscillatory dynamics) in order to function properly.

\subsection{Magnitude-Driven Latency Encoders}

When working in the physical world or in an information-rich environment, there can be a need to identify what information is relevant and to condition the output of the SNN based on this information. While methods that normalize values can be used to produce an output based on the intensity of the value, what if there is a need to produce a spike quickly in order to ``focus'' the network? Here, spikes that emit earlier in a train might be more important to process rather than ones that occur later on. 
In these cases, a different form of encoding -- what we will refer to as latency encoding -- can be used to produce spike timings that are driven and coordinated by the magnitude of the input. These spikes can be used to focus the network, bias its outputs, or to generally shift the dynamics such that spike timing is inversely proportional to stimulus intensity, i.e., higher magnitudes will result in earlier firing whereas lower magnitudes will cause a delay.

\subsubsection{Magnitude-Driven Timing}
\label{sec:magnitude_mechanics}

\begin{figure}[!t]
    \centering
    \includegraphics[width=0.70\linewidth]{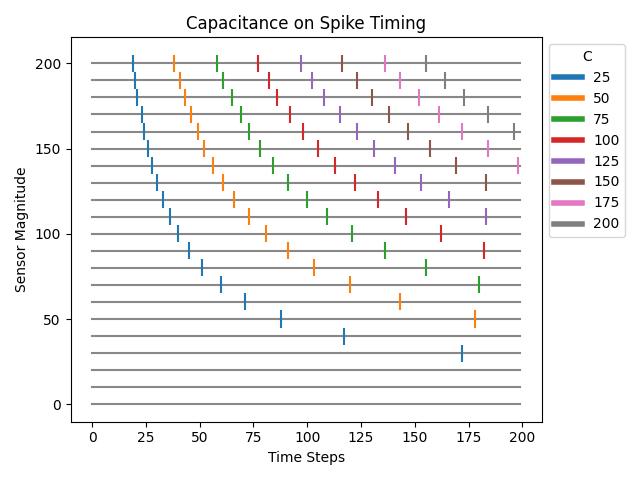}
    \caption{In this visualization, the effects of different dynamics within a latency encoder are observed when provided with constant input. Here, the resistance ($R$) is fixed at $R=1$ and the threshold value ($v_\text{thr}$) is held at $v_\text{thr}=15$. The different colored tick marks denote the time that each varied capacitance will produce a spike output for the given sensor magnitude (along the y-axis).}
    \label{fig:input:mag:constant}
\end{figure}

If the desired result is to produce a spike such that a larger input magnitude value results in the input unit spiking sooner, then treating the input sensory values as electrical current to a simplified resistor-capacitor (RC) circuit will model this desired behavior. 
Modeling the input as $i_{\text{in}}$ and the corresponding voltage $v$, at time $t$, we can then make use of the following mathematical model: 
\begin{equation}
    \label{eq:input:mag:base}
    v(t) = i_{\text{in}} R\big(1 - \exp(-t / RC)\big).
\end{equation}
Here, $R$ is the input resistance scaling factor and $C$ is the capacitance constant. When the circuit's input voltage value exceeds a threshold $v_\text{thr}$, a spike is emitted. In essence, this simplified RC circuit model means that, as the input grows larger in magnitude, the faster it will charge up the voltage $v(t)$ to its threshold (resulting in the emission of a discrete spike pulse). In practice, for this to work well, the input value must be treated as a constant. As such, the exact time for the $v(t) = v_\text{thr}$ can be calculated and thus the model can simply wait until the simulation time exceeds the computed time to then emit a spike. To solve for this time, we proceed as follows:
\begin{align}
    \nonumber v(t) &= i_{\text{in}} R\left(1 - \exp(-t / RC)\right) \\
    \nonumber v_\text{thr} &= i_{\text{in}} R\big(1 - \exp(-t /RC)\big) \\
    \nonumber \dfrac{v_\text{thr}}{i_{\text{in}} R} &= 1 - \exp(-t /RC) \\
    \nonumber \exp(-t / RC) &= 1 - \frac{v_\text{thr}}{i_{\text{in}} R} \\
    \nonumber \exp(-t / RC) &= \frac{i_\text{in} R - v_\text{thr}}{i_\text{in} R} \\
    \nonumber -t / RC &= \ln{\left(\frac{i_\text{in} R - v_\text{thr}}{i_\text{in} R}\right)} \\
    t &= RC\ln{\left(\frac{i_\text{in} R}{i_\text{in} R - v_\text{thr}}\right)}.
\end{align}
Note that further modifications are made to simplify this model down further. First, $R$ is traditionally set to $1$ and we then let $RC = \tau_\text{in}$. 
This allows for the timing calculation to be simplified to: 
\begin{equation}
    t = \tau_\text{in} \ln{\left(\frac{i_\text{in}}{i_\text{in} - v_\text{thr}}\right)}.
\end{equation} 
The behavior of this latency encoder (under constant input) is presented in Figure \ref{fig:input:mag:constant}.

Knowing the time $t$ when a fixed input will result in a spike can be very useful; however, sometimes the problem setting (e.g., in the case of a fixed input) makes obtaining this behavior difficult and thus other approaches are needed. If it is possible to assume that the input is only fixed for the window of time between time steps, it is possible to solve this problem via Euler integration in order to produce a time- and input-dependent version of this equation. 
Specifically, starting from the base equation, the first step is to solve for $dv(t) / dt$ through the following process: 
\begin{align}
    \nonumber v(t) &= i_{\text{in}} R\left(1 - \exp(-t /RC)\right) \\
    \nonumber \frac{dv(t)}{dt} &= i_\text{in}R \frac{1}{RC}\exp(-t / RC) \\
    \nonumber \frac{dv(t)}{dt} &= \frac{i_\text{in} R}{RC}\left(1 - \nonumber \frac{v(t)}{i_\text{in}R}\right) \\
    \nonumber \frac{dv(t)}{dt} &= \left(\frac{i_\text{in} }{RC} - \frac{v(t)}{RC}\right) \\
    \frac{dv(t)}{dt} &= \left(\frac{i_\text{in}R - v(t)}{RC}\right).
\end{align}
With the $dv(t) / dt$ obtained above, the equation for Euler integration can be used as follows: 
\begin{align}
    \label{eq:input:mag:continous}
    \nonumber v(t+\Delta t) &= v(t) + \Delta t \frac{dv(t)}{dt} \\
    \nonumber v(t+\Delta t) &= v(t) + \Delta t \left(\frac{i_\text{in}R - v(t)}{RC}\right) \\
    v(t+\Delta t) &= v(t) + \frac{\Delta t}{RC} \left(i_\text{in}R - v(t)\right).
\end{align}

\begin{figure}
    \centering
    \includegraphics[width=0.635\linewidth]{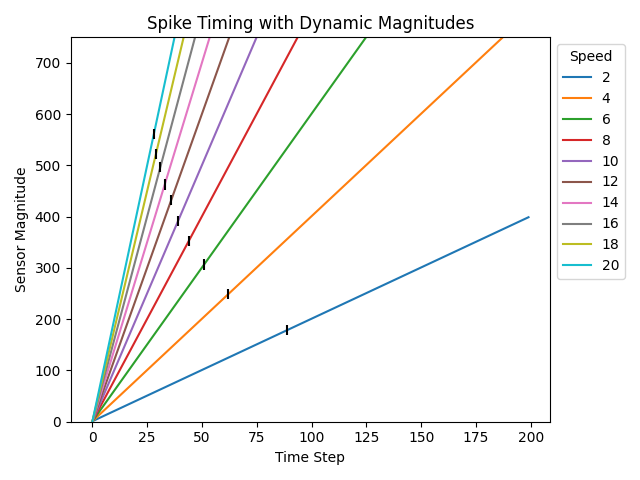}
    \caption{In this visualization, the effects of dynamic, varied input on a fixed encoder are depicted. Here, the resistance ($R$) is fixed at $R=1$, the capacitance ($C$) is fixed at $C=50$, and the threshold value ($v_\text{thr}$) is also fixed at $v_\text{thr}=15$. Each colored line represents a sensor reading that is increasing with time; the speed of this increase can be seen in the legend on the right. The black ticks indicate when this encoder would produce a spike output for each given sensor.}
    \label{fig:input:mag:dynamic}
\end{figure}

Concretely, the formulation of Equation (\ref{eq:input:mag:base}) found in Equation (\ref{eq:input:mag:continous}) can be used to calculate the spike emission of a latency encoding with a variable input (the behavior of which is visually depicted in Figure \ref{fig:input:mag:dynamic}); nevertheless, it is not always correct to use it. For instance, doing so will lose the clean approach for computing the expected firing time given a fixed input. As a result, the sensor and properties of the (sensory) input must be considered when choosing which formulation of latency encoding to use.

\subsection{Periodic Encoders}

There are times when the magnitude or intensity of the input data is less important than how that data changes over time. Specifically, if input data is periodic or wave-like in nature, then it be important to encode aspects unique to this kind of data, such as its frequency, while also maintaining/preserving intensity information, such as its amplitude. As a result, alternate methods of encoding information will be  needed.

\subsubsection{Phasor Encoding}
\label{sec:phasor_mechanics}

\begin{figure}
    \centering
    \begin{subfigure}[b]{0.5\textwidth}
    \includegraphics[width=1.0\linewidth]{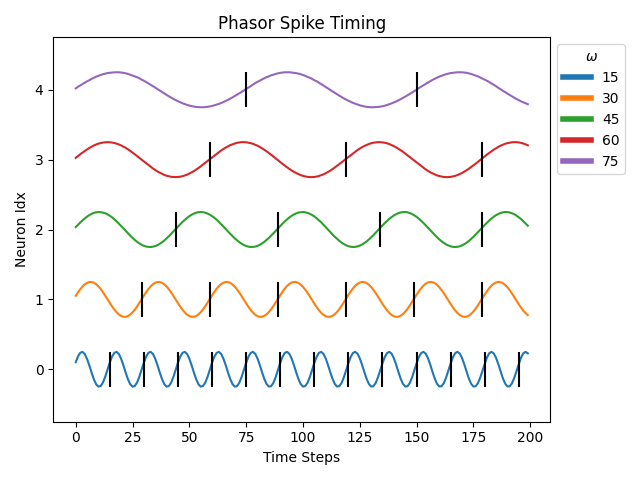}
    \end{subfigure}%
    \begin{subfigure}[b]{0.5\textwidth}
    \includegraphics[width=1.0\linewidth]{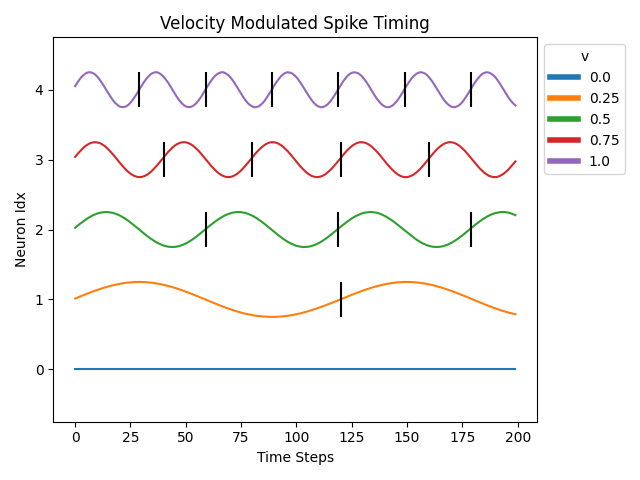}
    \end{subfigure}
    \caption{In this visualization, phasor-based encoding is depicted. 
    In the first sub-figure (\textbf{Left}), different firing frequencies ($\omega$) are presented while the input value is constant across all neuron. Each neuron's different induced spike patterns are superimposed on the graph.
    In the next sub-figure (\textbf{Right}), different inputs are used to modulate the velocities ($v$) of different identically initialized encoders. Once again each neuron's induced spike patterns are super imposed on the graph}
    \label{fig:input:periodic:phasor_omega}
\end{figure}

When working sensors that produce periodic data values that follow a simple sinusoidal pattern that is either known or approximated, then a phasor encoding scheme can be leveraged to produce more appropriate spike trains. 
In this setup, a phasor used to encode a sinusoidal input can be thought of as a spinning object where, upon every complete rotation of the object, a spike is then emitted. In general, if $f_e$ is the expected frequency of events (usually set to match the input signal's frequency) the angular speed ($v$) of the phasor can be calculated in the following manner: 
\begin{equation}
    v = \frac{2\pi \Delta t}{1000}f_e
\end{equation}
where $f_e$ is measured in \textit{events per second} and the units of $v$ work out to be \textit{radians per timestep}. From here, at every time step, the speed is then added to the current angle of the phasor and, if the phasor passes through a multiple of $2 \pi$, then a spike is emitted. Note that the phase offset can be applied by setting the current angle of the phasor prior to starting the simulation. See Figure \ref{fig:input:periodic:phasor_omega} for a demonstration of the phasor-based encoder. 

Notably, there are modifications that can be made to the above in order to produce variations within the encoded spike train. 
The first of these is to inject noise into the update of the angle at every timestep. Given the update $v$, a small delta sampled from a uniform distribution (centered around zero) can be added to produce a variation in the expected angle. These small deltas over time will average out to the expected frequency of events but have the potential to shift the emitted spike timing by one or two timesteps before or after the expected output. 
The second modification that can be made to produce variations is to couple the phasor with an intensity measure. From here, the intensity measure can be used to adjust the angular phasor to produce spikes at a modulated frequency. This modulation can happen online so as to adjust the frequency of spikes with the coupled intensity measurement, while still producing periodic spikes.


\subsubsection{Wave Encoding}
\label{sec:wave_encoder_mechanics}

\begin{figure}[!t]
    \centering
    \includegraphics[width=0.75\linewidth]{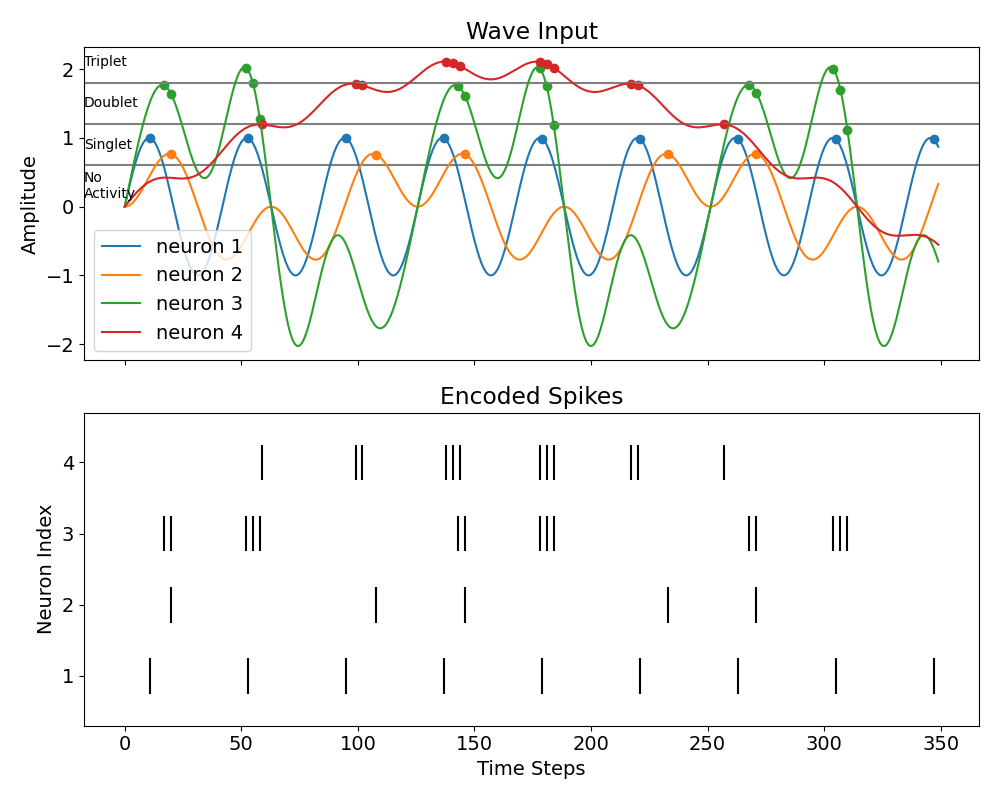}
    \caption{Here, we depict four different waves (top-half diagram) that are parsed by the same wave encoder in order to produce different spike trains (bottom-half diagram). 
    The top-half diagram shows not only the different waves but also highlights where the encoded spikes are relative to the maxima that produced the event. The labeling on the left also shows the different windows where event will produce no activity, a singlet, doublet, or a triplet. 
    The bottom-half diagram shows a raster plot of the same encoded events but in a fashion that more readily highlights the periodic nature of the events themselves. For this figure/demonstration, an inter-spike refractory period of $2$ (ms) was chosen and a ``cost-per-spike'' $c$ of $0.6$ was selected.}
    \label{fig:input:periodic:wave}
\end{figure}

The most general form of encoding periodic sensor data involves detecting local maxima in the raw signal and then generating spikes, accordingly. This method offers the advantage of no longer requiring a fixed frequency or predefined waveform characteristics (aside from  requiring the signal to oscillate locally). Fundamentally, this encoding strategy tracks two essential features: 
\textit{(1)} whether the sensor reading $x(t)$ is currently increasing, and 
\textit{(2)} the most recent value that has been observed. 
When the sensor reading was previously increasing yet its current value has fallen below (less than) the prior observation, a local maximum is identified, and a spike is subsequently generated. 
In its minimal format, this process is sufficient to produce a discrete spike train from a continuous stream of sensor data values. The general form for this scheme, if there is a spike ($S$) at a specific time ($t$), is as follows: 
\begin{equation}
    S(t) = \begin{cases}
    1 & \text{if } x(t-2) < x(t-1) \wedge x(t-1) > x(t) \\
    0 & \text{otherwise}.
    \end{cases}
\end{equation}
However, the above approach discards a critical dimension of information; specifically, it discard the amplitude of the waveform at the time that a spike is emitted. To retain this amplitude-dependent information, the encoder must instead interpret each local maximum as an event that correlates to a burst of spikes, where the number of spikes is modulated by the instantaneous amplitude. As a result, low-amplitude signals result in weaker stimulation than high-amplitude signals, even when they follow an identical event structure. 
A general equation for calculating the number of spikes ($n_{\text{spikes}}$) that a specific event will yield can be written as: 
\begin{equation}
    n_{\text{spikes}} = \left\lfloor \frac{A}{c} \right\rfloor
\end{equation}
where \(A\) is the amplitude of the waveform at the time that the event began and \(c\) is a ``cost-per-spike.'' In Figure \ref{fig:input:periodic:wave}, we visualize the behavior of a wave spike encoder. 
We remark that if the amplitude at the time of an event is not large enough then no spikes will be emitted, even though an event has occurred. Moreover, if the amplitude does not reach the threshold for the next spike (i.e., the next integer multiple of the cost-per-spike \( c \)), any excess amplitude below that threshold will be effectively lost.

Finally, there is one more point that must be addressed when working with burstiness -- inter-spike refractory times. Generally, successive spikes with a refractory period spacing them apart (i.e., neurons fatigue after firing and are not immediately able to re-fire as easily in the near future) are commonly observed in neurobiological systems. 
As a result, a refractory time between spikes in a burst of spikes is chosen to be smaller than the inter-event refractory time yet still large enough such that each spike within the ``burst'' does not bleed into the previous or next one.

\subsection{Event Encoders}
In this work, the final method of encoding sensory data that we will cover relates to either just reading in event streams or converting temporal information into event(-like) streams. 
Unlike the previous encoders that produce binary pulse streams of either a spike or no spike, an event encoder produces one of three possible discrete output signals $\{-1, 0, 1\}$; these outputs correspond to a `decrease event', `no event', or an `increase event'. Event encoders are a bit different from the other encoders mentioned in this article given that many of them operate at a hardware level and, when implemented in software, these types of encoders are often meant to simulate what the hardware would produce naturally. 

One of increasingly popular sensor that has been utilized to produce event streams is an event camera or dynamic vision sensor (DVS) \cite{prophesee2023dvs}; note that these sensors detect changes in incoming light on a pixel-by-pixel basis to produce either a decrease or increase event based on the direction of the change itself. Note that these pixels do not output a `no-op' or `no event' when there is no change; instead, they just remain silent which means that, to truly take full advantage of these sensors, the information processing performed by the entire model itself (i.e., the SNN) needs to also be able to be handled asynchronously, which usually requires specialized hardware 
(such specialized hardware includes neuromorphic chips, e.g., \cite{davies2021advancing}, which will not be specifically discussed in this work). 
Even without access to specialized hardware, it is often rather useful to simulate the behavior of the event-streams produced by DVS as a preliminary experimental step in validating/testing SNN models that are to be deployed for low-power (edge computing)  applications. 

\subsubsection{Simulating Vision Sensor Event Streams}
\label{sec:event_streams}

In order to simulate event streams, utilizing a dataset as the ``environment'', it is important to remember that events correspond to \emph{changes} within the data stream. This means that, if the dataset is a collection of still images or constant sensor readings, then technically no events would ever be produced (yielding an empty stream). As a result, a vision sensor would be more appropriately simulated if the data exhibits some form of ``flow''; in other words, video frames or time series data points would fit this criterion well given that the information these kinds of data contain can be described in terms of an inherent time-axis (or sequential ordering). 

\begin{figure}
    \centering
    \includegraphics[width=\linewidth]{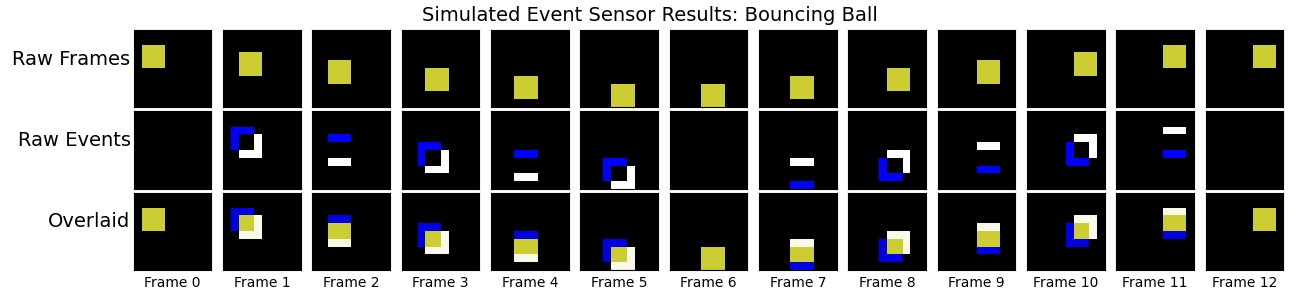}
    \caption{The above depicts how the frames (the first, \textbf{top} row) for a ``bouncing'' square in the viewport would be transformed into a vision sensor-like event stream. 
    In this example, at each time step, the yellow square moves either up or down and, at every other frame, it moves one step to the right. 
    In the second (\textbf{middle}) row, the increasing and decreasing events are displayed in white and blue, respectively. 
    It is important to note that once the square reaches the top or bottom of its ``bounce'' it is paused for a single frame (i.e., at frames $6$ and $12$); this can be seen reflected in the event data by the empty frames present in the raw events row. 
    The third/final (\textbf{bottom}) row depicts the same event data overlaid over the raw video frames; here one may observe that the leading and trialing edge of the square is reflected in the events at each time step.}
    \label{fig:event_stream_sim}
\end{figure}

The process to simulate the events, given data embodying a temporal ordering, entails a rather straightforward approach. Starting from the first frame (assuming video data), one simply needs to compare the difference between the current frame and the previous one so as to produce a delta/difference in intensity (values). 
From here, this delta is passed through bound checking routine in order to verify that the magnitude of the delta is sufficiently large. A discrete event is emitted based on the sign of the computed delta. Formally, the delta (assuming discrete time-steps for the series of sensory frames) is calculated in the following manner:
\begin{equation}
    \delta_I(t) = I(t) - I(t-1) 
\end{equation}
where $I(t)$ is the intensity of the sensor at time $t$. In order to compute if an event $E(t)$ is emitted at time $t$ (as well as the type of event), we perform the following calculation: 
\begin{equation}
    E(t) = \begin{cases}
    -1 & \text{ if } |\delta_I(t)| > \epsilon_l \wedge \delta_I(t) < 0 \\
    1 & \text{ if } |\delta_I(t)| > e_h \wedge \delta_I(t) > 0 \\
    0 & \text{ otherwise}
    \end{cases}
\end{equation}
where $\epsilon_l$ and $\epsilon_h$ are the configured thresholds for the decreasing and increasing events, respectively, of which the delta needs to exceed in order to produce the relevant event (type). Note that the event $E(t)$ is signed to mark whether it is a decreasing (negative) or increasing (positive) event. 
In Figure \ref{fig:event_stream_sim}, we visually demonstrate how such a vision sensor event stream is generated given video frame information.  

\noindent 
\textbf{Limitations:} While this method does work for producing events and can even be used to train preliminary SNN models on event streams, there are some stark differences between certain sensors and these simulated event pulses. For instance, when using the Prophesee DVS \cite{prophesee2023dvs}, if there is a light emitter in the frame, such as an LED, a blooming effect will be produced that is not easily captured by  traditional cameras. 
Furthermore, there is often much more noise in real-world event streams as compared to synthetic ones. Some of this noise can be simulated by applying centered Gaussian noise over an image in order to include random events. This noise works decently well at creating additional meaningless events to the stream; however, determining the correct amount of noise is not always straightforward. In addition, such noise will not capture and replicate the full spectrum of the real-world's generated ``extra events'', as these are not purely random and likely the consequence of other unobserved processes.

\subsubsection{Processing Events}
\label{sec:processing_events}

SNN models that are trained with data that comes from event stream sensors must further be able to handle their asynchronous nature. 
The most biological approach to this would be to process each and every event that arrive in parallel, having each input neuron in the model designed such that they can be stimulated individually and pass each encountered event into the (internal layers of the) system. This setup/design, unfortunately, quickly becomes feasible for models that are very large and often run on traditional von Neumann architectures. 
It is, however, possible to instantiate and run these on specialized hardware platforms, including field-programmable gate arrays (FPGAs) and neuromorphic chips, which are capable of handling this type of data stream due to their ability to process in-memory \cite{mead2002neuromorphic,ororbia2023mortal}. 

While specialized hardware exists for instantiating SNNs to process direct event streams, many research projects face not only the hurdle of affording these platforms but also the difficulty of porting/translating their software models accurately, faithfully, and effectively to the hardware (or a hardware specification that lends itself to instantiation on such platforms).  
Nevertheless, to aid in speeding up the processing of event streams in non-speciality hardware, there do exist useful methods for (pre-)processing the event stream. The primary scheme entailing the `binning' of the input stream. Under this scheme, the input events can be collected over a window of time and then compressed o aggregated into a single time-step of data; this allows for that singular time-step to be processed like any intensity-based input to the system. 
Whether or not one is using specialized hardware or a binning approach to process events, there is one more notable issue to clarify when working with events and SNNs -- since SNNs operate with non-signed binary signals, the trinary pulse values (i.e., $\{-1, 0, 1\}$) that come from DVS streams will need to be converted appropriately. This entails creating a mapping layer where each DVS trinary unit will have two out-going connections -- one for increasing events and another one for decreasing ones. This structural setup will not only facilitate downstream processing via an SNN but also allow it to effectively infer and learn from both the leading and trailing edge of objects containing within a camera image without them bleeding into one another.

\section{Neuronal Models}
Once an input encoding scheme has been decided on and designed / integrated properly, the next step in designing an SNN model to simulate is to consider the neuronal units that will constitute its hidden (and possibly output) populations or layers. However, there is a plethora of neuronal models to make use of when undertaking this design, ranging from complex, slow-to-simulate bio-physically accurate models \cite{hodgkin1952quantitative} to vastly under-simplified yet fast discrete-emission units \cite{mcculloch1943logical}. In addition, one could also consider whether to emulate a neuronal model and its spike-emission function via iterative integration of differential equations or through the design and use of spike response kernels \cite{gerstner1998spiking,gerstner2002spiking}; we will narrow our focus on the integration of differential equations used to describe the internal components (and their interactions) of neuronal spike models. 
In this work, we will focus on two important, yet approachable, building blocks that often serve as starting points for SNN circuit design and modeling -- the leaky integrate-and-fire and the resonant-and-fire computational models. These two types of neuronal models are not only approachable for newcomers to SNN design and operation but also entail faster simulation speed without simplifying biological neural mechanics and dynamics too much (although they will not capture much of the nonlinear, neuro-anatomically faithful behavior of complex models such as the Hodgkin-Huxley model \cite{hodgkin1952quantitative} or multi-element emulations \cite{izhikevich2003simple}). 

\subsection{Leaky Integrate-and-Fire Neurons}

In mathematics, the leaky integrator is a specific form of a linear ordinary differential equation (ODE)\footnote{In general, the form of the leaky integrator ODE for modeling an entity $x$ is $dx/dt = -a x + b$, where $a$ is the input leakage rate and $b$ is the received input at time $t$.}, often utilized to characterize a system that carries out an integration process of its input values yet gradually loses (or ``leaks'') some small amount of this input over time. In computational neuroscience, the leaky integrator serves as a useful building block in describing the dynamics of a neuronal cell and, when it is combined with some mechanism for emitting action potentials once it crosses a particular critical value, we arrive at the well-known leaky integrate-and-fire (LIF) spiking neuron \cite{burkitt2006review,naud2012performance}. Historically, the equations that characteristically described the LIF were first proposed by Kenneth Stein \cite{stein1965theoretical} in $1965$, offering an early, useful simplified mathematical model of how a neuron's membrane potential behaves as a function of time. 

An LIF, as well as general integrate-and-fire-like model generations \cite{brette2005adaptive,teeter2018generalized}, essentially requires at least two key elements to be specified: 
\textbf{1)} at least one equation to describe how its membrane potential evolves with time, and 
\textbf{2)} a function/mechanism to produce action potentials/spikes (often in relation to the properties of the membrane potential). 
Crucially, the LIF makes no attempt to describe the form or shape of the action potentials that it emits, i.e., it assumes that the action potentials of a neuron always have approximately the same shape, which means that an LIF pulse's shape will carry no useful information and, instead, it is the absence or presence of a pulse that contains information. 

In essence, the LIF entails that a neuron $i$'s built-up voltage $v_i(t)$ (note that, later on, when discussing a single neuron, we will sometimes drop the index $i$ to reduce notational clutter), which contains the integrated effect of all of its inputs (or all the result of all of the inputs that are charging up the neuron), reaches a particular threshold from below, we describe that neuron as firing a pulse or spike; the moment that this threshold is crossed or breached, it is time-stamped by the firing time $t^f_i$. Action potentials produced by an LIF can then be viewed simply as the ``events'' that occur at precise moments in time.

\subsubsection{Numerical Specification of the LIF}
\label{sec:lif_specification}

The LIF neuron, depending on the details of the model one wants to emulate, consists of either one or two ordinary ODEs. The key difference between the single and dual-ODE models is if the current input to the neuron is modeled as a single point or if it must follow the dynamics of an ODE as well. 

To start, we will first treat point-wise electrical current and then, later, develop the extension for modifying the equation treat it as time-varying current flow. 
For these models, we will let $j(t)$ denote the current at given time $t$ (which itself can further be a function of directly injected current $i(t)$, which could be provided externally or via the aggregation of signals/pulses emitted by other other neurons); likewise, we will let $v(t)$ denote the voltage at that same time. As these are the only dynamical parts of the system, all other coefficients will be assumed to be constant for the duration of the model. The equation that describes the dynamics of $v$ over time is as follows: 
\begin{align}
    \label{eqn:lif_voltage}
    \tau_v \frac{d v(t)}{d t} = -\gamma_v \big(v(t) - v_{\text{rest}}\big) + \rho_v j(t)
\end{align}
where $\tau_v$ is the membrane potential time constant, $\gamma_v$ is a leak coefficient, and $\rho_v$ is the neuronal resistance. The ODE in this form is usable but not easily implemented; as a result, it is common practice to solve this ODE over a fixed $dt$. Since we are letting $j(t)$ be a point-wise current ($j(t) = i(t)$), it is assumed that it operates/behaves as a step function where $j(t)$ is held constant over the time window. Likewise, since this will be solved via Euler integration, $v(t)$ will be equal to $v(t_0)$ for the entire window of time as well. 
The derivation for $v(t+\Delta t)$ then proceeds as follows: 
\begin{align}
    \nonumber \tau_v \frac{d v(t)}{dt} &= -\gamma_v \big(v(t) - v_{\text{rest}}\big) + \rho_vj(t) \\
    \nonumber d v(t) &= \frac{1}{\tau_v}\big(-\gamma_v \big(v(t_0) - v_{\text{rest}}\big) + \rho_vj(t_0)\big)dt \\
    \nonumber \int_{v(t_0)}^{v(t_0 + \Delta t)} dv &= \int_{t_0}^{t_0 + \Delta t}\frac{1}{\tau_v}\big(-\gamma_v \big(v(t_0) - v_{\text{rest}}\big) + \rho_vj(t_0)\big)dt \\
    \nonumber v |_{v(t_0)}^{v(t_0 + \Delta t)} &= \frac{1}{\tau_v}\big(-\gamma_v \big(v(t_0) - v_{\text{rest}}\big) + \rho_vj(t_0)\big) * t |_{t_0}^{t_0 + \Delta t} \\
    \nonumber v(t_0 + \Delta t) - v(t_0) &= (t_0 + \Delta t - t_0) * \frac{1}{\tau_v}\big(-\gamma_v \big(v(t_0) - v_{\text{rest}}\big) + \rho_vj(t_0)\big) \\
    v(t_0 + \Delta t) &= v(t_0) + \frac{\Delta t}{\tau_v}\big(-\gamma_v \big(v(t_0) - v_{\text{rest}}\big) + \rho_vj(t_0)\big). \label{eqn:lif_euler_form}
\end{align}
Now that there is a clean way to calculate the new voltage at every time-step, we can move on to the next step in the dynamics of the LIF -- depolarization (where a neuron's ion channels open and let positive ions flow into the cell, increasing its membrane to voltage to cross its threshold value). 
Depolarization through thresholding is the general method-of-choice for computing the spike pulses emitted by an LIF neuron. The computation of a spike event emission takes the form of a piece-wise function using the newly calculated voltage (from the Euler-integrated ODE of Equation \ref{eqn:lif_euler_form}) as follows: 
\begin{align}
    s(t) = 
    \begin{cases}
        1 & v(t) > \theta(t) \\
        0 & \text{otherwise}
    \end{cases} \label{eqn:spike_emission}
\end{align}
where $\theta(t)$ is the voltage threshold of the neuron. 
Once the voltage value $v(t)$ of the neuron crosses this threshold, it will then be set directly to its rest voltage value $v_{\text{rest}}$ (i.e., repolarization) and sometimes to a state lower than its resting-state voltage (i.e., hyperpolarization) where $v_{\text{reset}} < v_{\text{rest}}$. 
This post-spike mechanism is implemented as follows: 
\begin{align}
    v(t) = 
    \begin{cases}
        v_{\text{reset}} & s(t) = 1 \wedge \text{the neuron is to be hyperpolarized} \\
        v_{\text{rest}} & s(t) = 1 \wedge \text{the neuron is to be repolarized}.
    \end{cases} \label{eqn:post_spike_setting}
\end{align}

The threshold value $\theta(t)$ for spike emission can be fixed across time or configured to be another dynamical system, similar to the voltage. If a dynamic value is desired for the threshold, there are several ways to design this form of the threshold -- notably, one can consider an isolated/independent, per-neuron threshold or thresholds that are functions of a set or population of neurons. 
First, we will examine the isolated per-neuron format; in this case, we will label $\theta_{\text{base}}$ as the base value for the threshold and $\hat\theta(t)$ as the shift from this base value (which is initially set to zero). 
Then, at every time-step, $\hat\theta$ will be adjusted in accordance to the following ODE dynamics (note that this is also a leaky integrator):
\begin{equation}
    \frac{d \hat{\theta}(t)}{d t} = -\frac{1}{\tau_\theta}\hat{\theta}(t) + \kappa_\theta s(t)
\end{equation}
where $\tau_\theta$ is the homeostatic variable time constant and $\kappa_\theta$ is an increment value. The actual threshold used to compare the voltage $v(t)$ against is, in effect, the sum of the dynamic component and the base threshold value. This leads the final dynamic threshold equation formally presented below: 
\begin{equation}
    \theta(t) = \theta_\text{base} + \hat\theta(t). \label{eqn:per_neuron_threshold}
\end{equation}
While this method will work for a neuron in isolation, there might be the design need  for a population of neurons to exhibit specific behaviors collectively (e.g., strongly enforcing degrees of sparsity in the population's activity overall). In this case, both sharing and adjusting a threshold as a function of the neural population's behavior can be more beneficial. Commonly, some desired effect -- such as having $n$ events happen per time-step or to only have a small proportion of neuronal cells emit at any given instant -- is desired. To achieve more collective/aggregate effects of this kind, we still use the same base value and shift value developed earlier (as in Equation \ref{eqn:per_neuron_threshold}); however, instead we will now employ these mechanisms for a specific group or population of neurons which we will denote with the superscript $\ell$ (i.e., the $\ell$-th population layer in an SNN circuit structure). 
The equation used to govern the population-driven $\hat\theta$ for this formulation is then as follows: 
\begin{equation}
    \frac{d \hat{\theta}^\ell(t)}{d t} = \kappa_\theta \big(n - \sum_{i=1}^ks^\ell_i(t)\big) \label{eqn:population_driven_threshold}
\end{equation}
where $n$ is the number of expected events, $\kappa_\theta$ is again a variable increment value, and $k$ denotes the number of neurons within the $\ell$-th population. Note that Equation \ref{eqn:population_driven_threshold} can lead to very sparse behavior within a group of spiking neuronal cells (the spike ``codes'' become quite sparse). 

When combining Equations \ref{eqn:lif_euler_form} and \ref{eqn:spike_emission} (with, optionally, either Equations \ref{eqn:per_neuron_threshold} or \ref{eqn:population_driven_threshold}), the complete dynamics of an LIF have been fully specified for the case of incoming point-wise electrical current $j(t)$.
Nevertheless, in biology, current does not exist as a point-wise step function; instead it can be better likened to behaving as dynamical system itself, similar to that of the one we used to model the membrane potential. 
In fact, the base equation to be used to model electrical current $j(t)$ will look practically the same as that used for $v(t)$ (i.e., it will follow leaky integrator dynamics), as in below: 
\begin{align}
    \tau_j \frac{d j(t)}{d t} = -\gamma_j v(j) + \rho_ji(t). \label{eqn:dynamical_current}
\end{align}
Here $\tau$, $\gamma$, $\rho$ are functionally identical to the coefficients used in the voltage ODE (Equation \ref{eqn:lif_voltage}). $i(t)$ is the signal that was transmitted at time $t$. This means that the the current at any given time $t$ will work out the same as it did for the voltage and produce: 
\begin{align}
    j(t_0 + \Delta t) &= j(t_0) + \frac{\Delta t}{\tau_j}\big(-\gamma_j j(t_0) + \rho_ji(t_0)\big). \label{eqn:current_euler_form}
\end{align}
Note, in an SNN implementation, the above Euler-integrated ODE for the current $j(t)$ will be calculated prior to the the membrane potential equations for voltage $v(t)$. 
One pragmatic issue when working with SNNs is related to whether or not enough electrical current has been injected into Equation \ref{eqn:lif_voltage} in order to obtain at least a few spikes (as, otherwise, the SNN would be entirely quiescent); if the modeler uses Equation \ref{eqn:current_euler_form} or more complex formats (that model synaptic conductance channels, e.g. $\alpha$-amino-3-hydroxy-5-methyl-4-
isoxazolepropionic acid AMPA, $\gamma$-minobutyric acid GABA$_{A}$, as in \cite{zenke2015diverse}; this would change Equation \ref{eqn:lif_voltage} into a multi-channel form that models multiplicative reverse potentials) then electrical current is integrated/decays over time and thus an LIF is more likely to spike even under sparser input stimuli conditions. However, if the instantaneous form is used -- $j(t) = i(t)$ -- it a practical configuration of Equation \ref{eqn:lif_voltage} to ensure that spikes are more likely is to set $\rho_v = \tau_v/dt$, which will effectively amplify the current enough such that it roughly mimics the effect of an ODE-based $j(t)$ (especially if further employed alongside a subtractive threshold, as described in Section \ref{sec:time_discretization_error}). 

Ultimately, the full LIF neuronal model is then simulated, at each time-step under an integration time constant $\Delta t$, through the following process: 
\begin{enumerate}[label=(\arabic*),noitemsep,nolistsep]
    \item calculate the injected current $i(t_0 + \Delta t)$;
    \item calculate $j(t_0 + \Delta t)$ as either $j(t_0 + \Delta t) = i(t_0 + \Delta t)$ (point-wise current) or via Equation \ref{eqn:current_euler_form} (dynamical current flow); 
    \item calculate the membrane potential / voltage $v(t_0 + \Delta t)$ via Equation \ref{eqn:lif_euler_form}; and, 
    \item compute the spike event/pulse $s(t)$ via Equation \ref{eqn:spike_emission} and hyper/repolarize neuron via Equation \ref{eqn:post_spike_setting} and then move time  forward ($t \leftarrow t + \Delta t$) and return to (1).
\end{enumerate}

\paragraph{Biological Starting Point:} 
From a neurobiological point-of-view, and as a result of experimental studies, there are generally some typical values one can set for many of the constants in the differential equations developed above. 
Specifically, for the LIF membrane potential ODE (Equation \ref{eqn:lif_voltage}), it is more common to see a leak rate of $\gamma_v = 1$ and a resting potential or baseline voltage (for a neuron that is not firing) $v_{\text{rest}}$ that is within the range of $[-73, -65]$ millivolts (mV); 
if a voltage reset value $v_{\text{reset}}$ is employed to model hyperpolarization, it is set to be lower than the threshold (and the resting potential) with values in the range of $[-75, -69]$ mV. For the threshold itself, a base threshold value $\theta_{\text{base}}$ for emitting an action potential is sometimes set to value in the range of $[-55,-50]$ mV, e.g., a reasonable configuration would be $\theta_{\text{base}} = -54$ mV. 
The LIF's resistance $\rho_v$ can be set within the range of $100$s of Ohms ($\Omega$) to several megaOhms (M$\Omega$). 

The time constant is usually decided based on the time-course that the modeler or experimentalist wants to consider, though generally these are set on the temporal order/scale of milliseconds (ms) or microseconds ($\mu$s), e.g., depending on the order/scale, $\tau_v$ could be set within the range of $10$-$100$ ms (note that this is the amount of time for an LIF neuron's membrane potential to reach $63.2$\% of its final value after a step input). 
The refractory period, if one is used, is typically set to a few milliseconds of simulated time. What values should be set for mechanisms such as the adaptive threshold are typically less clear. 
Values for the integration time constant $dt$ that is used to carry out simulation are typically on the order of milliseconds as well (e.g., $0.1$-$1$ ms) and vary depending on the accuracy desired from the neuronal model to be simulated. 

\subsubsection{Tuning LIF Dynamics}
\label{sec:tuning_lifs}

Given our formulation above, we will next examine the effects that the membrane time constant, decay, and input coefficients have on LIF dynamics. In particular, we will simulate the LIF's behavior under specific settings/configurations in order to understand its behavior and overall operation.

\begin{figure}[!ht]
    \centering
    \includegraphics[width=0.6\linewidth]{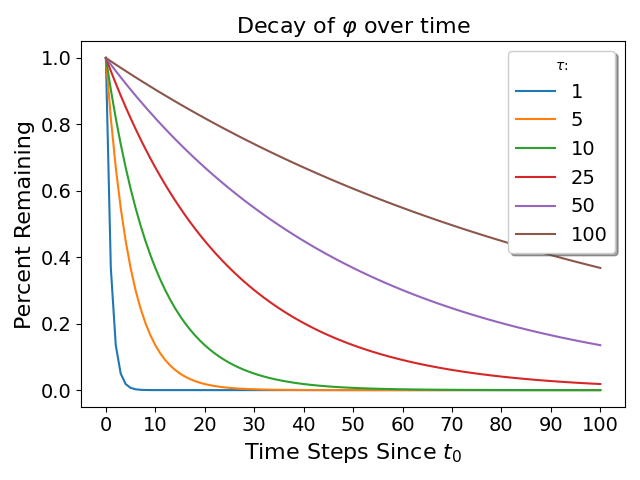}
    \caption{
    Comparison of different time constants ($\tau$; on the order of milliseconds) and their effect on the decay rate for exponentially decaying values in the context of the leaky integrate-and-fire (LIF) neuronal model.
    }
    \label{fig:phi_decay}
\end{figure}

\noindent 
\textbf{The Membrane Time Constant: }
To gather an intuition behind membrane time constants -- and time constants in general -- in relation to the ODEs commonly found in computational neuroscience, we will first look at a general equation using abstract terms. If there is a value ($\varphi$) that exponentially decays with time ($t$), the dynamics of that decay can be adjusted with time constant $\tau$. Starting with the equation: 
\begin{equation}
    \label{eq:tau_general}
    \tau_\varphi \dfrac{\partial  \varphi(t)}{\partial t} = -\varphi(t)
\end{equation}
we can then solve for the value exact value of $\varphi$ given the current time and the value at some time $t_0$. This results in the following derivation:
\begin{align}
    \label{eq:tau_general_solved}
    \nonumber \tau_\varphi \dfrac{\partial \varphi(t)}{\partial t} &= -\varphi(t) \\
    \nonumber \int \dfrac{1}{\varphi(t)} \partial \varphi(t) &= \dfrac{-1}{\tau_\varphi} \int \partial t \\
    \nonumber \ln(\varphi(t)) &= \dfrac{-1}{\tau_\varphi}(t + C) \\
    \nonumber \varphi(t) &= C e^{\dfrac{-t}{\tau_\varphi}} \\
    \nonumber \varphi(t_0) &= C e^{\dfrac{-t_0}{\tau_\varphi}} \\
    \nonumber C &= \varphi(t_0) e^{\dfrac{t_0}{\tau_\varphi}} \\
    \varphi(t) &= \varphi(t_0) e^{\dfrac{t_0-t}{\tau_\varphi}}.
\end{align}
This general form for exponentially decaying values is used in a few places throughout the LIF; such as in the voltage and current ODEs. Note that, in the LIF, the time constant $\tau_\phi$ (i.e., $\tau_v$ o $\tau_j$ in the LIF dynamics) is not the only coefficient governing the dynamics of its equations; as a result, we will treat them in what follows. 

\noindent 
\textbf{The Decay and Input Coefficients: }
In addition to the time constant $\tau_v$ found in Equation \ref{eqn:lif_voltage}, there are coefficients that scale both the decay rate ($\gamma_v$) and the incoming input ($\rho_v$). These two interact with $\tau_v$ to either strengthen or weaken the effects of the time constant. By setting the resting potential to $v_{\text{rest}} = 0$, starting with:
\begin{equation}
    \label{eq:coeff_1}
    \tau_v \frac{\partial v(t)}{\partial t} = -\gamma_v v(t) + \rho_v j(t)
\end{equation}
we can then say that if $\gamma_v = 1$ and $j(t) = 0$ for all $t$, then this equation functions identically to the one defined in Equation \ref{eq:tau_general_solved}. 
Now as $\gamma_v$ changes, its effects on $\tau_v$ will be inversely proportional to the change. The effects of scaling $\tau_v$ and the decay can be seen in Figure \ref{fig:phi_decay}. An alternative method for setting $\gamma$ and $\tau_v$ is to choose how many time-steps a neuron should take before it is set to a specified proportion of the original value, given no extra input. As an example, if a neuron is to reach $10\%$ of its original voltage -- or $0.1v(t_0)$ -- within a specific $\Delta t$, we may solve for the ratio via the following steps: 
\begin{align}
    \label{eq:lif_ratio}
    \nonumber \tau_v \frac{\partial v(t)}{\partial t} &= -\gamma_v v(t) \\
    \nonumber v(t) &= v(t_0) \exp((\gamma_v /\tau_v)(t_0 - t)) \\
    \nonumber v(t_0 + \Delta t) &= v(t_0) \exp((\gamma_v / \tau_v)(t_0 - (t_0 + \Delta t))) \\
    \nonumber 0.1v(t_0) &= v(t_0) \exp(- \gamma_v \Delta t / \tau_v) \\
    \nonumber 0.1 &= \exp(- \gamma_v \Delta t / \tau_v) \\
    \dfrac{-\ln(0.1)}{\Delta t} &\approx \dfrac{2.3}{\Delta t}= \dfrac{\gamma_v}{\tau_v}.
\end{align} 
Beyond scaling $\gamma_v$, the resistance value ($\rho_v$) in front of $j(t)$ can be further used to control the impact that the injected/incoming current $j(t)$ supplied to the LIF neuronal system has. Traditionally, $\rho_v$ is scaled in terms of $\tau_v$ or $(\rho_v = \tau_v \rho_v')$ such that a consistent amount of current will be provided regardless of the LIF system's decaying dynamics.

\subsection{Time Discretization Error and Reset by Subtraction}
\label{sec:time_discretization_error}

Several popular SNN frameworks employ time-based simulation with a fixed time-step $\Delta t$.  Under these conditions, it is critical to consider the potential for information loss due to discretization error. 
For example, consider a simplified LIF with a linear leak factor and a constant stimulus such that its membrane charge rate is $r$.  In this case, the time required to charge the membrane to its threshold is $t=\theta/r$. However, the simulation (of this LIF) will produce a spike at $t'=\lceil t/\Delta t\rceil\Delta t=t+\epsilon$. Therefore, the difference in the number of spikes that occur with and without discretization within a period $T$ is approximately: 
\begin{equation}
\Delta n = \left\lfloor \frac{T}{\theta/r} \right\rfloor - \left\lfloor \frac{T}{\lceil \theta/r\Delta t\rceil\Delta t} \right\rfloor\approx \frac{Tr^{2}\epsilon}{\theta(\theta-r\epsilon)} . 
\label{eqn:nerror}
\end{equation}
The approximation on the right-hand side of (\ref{eqn:nerror}) is made by removing the floor functions. Note that this will only produce an approximation error of $1$, at most.  We can also estimate the spike rate error due to quantization as follows: 
\begin{equation}
\Delta \nu \approx \frac{r^{2}\epsilon}{\theta(\theta-r\epsilon)} . 
\label{eqn:rateerror}
\end{equation}

Crucially, from (\ref{eqn:nerror}) and (\ref{eqn:rateerror}), we see that there  significant loss of information occurs when the time-step is too large, the threshold voltages are too small, or the neuron charge rate is very high.  
At the core of the problem is the fact that the information represented by the membrane potential's charge, beyond $\theta$ (from $t$ to $t'$), is lost when the neuron fires at $t'$ and the membrane is reset to a constant (e.g. zero) voltage. One widely-used approach to address this issue is to simply carry over the superthreshold charge to the next charge cycle.  This can be done by subtracting the threshold from the membrane potential when a neuron fires instead of resetting it to a constant value \cite{rueckauer2016theory}; this means modification of Equation \ref{eqn:lif_voltage} to incorporate the subtraction of the voltage (instead of using Equation \ref{eqn:post_spike_setting}, post-Euler update). 
Now, if a neuron spikes $n'$ times during a discretized period $T$, then by the end of the period, the potential will be $v(T)=Tr-\theta n'$, leading to the following: 
\begin{equation}
\Delta n \approx \frac{T}{\theta/r}-\frac{Tr-v(T)}{\theta}=\frac{v(T)}{\theta}
\end{equation}
\begin{equation}
\Delta \nu \approx\frac{v(T)}{T\theta}
\end{equation}
where, again, the approximations stem from the removal of the floor function in counting the number of spikes within period $T$. Notice now, desirably, that the error has been significantly improved. Furthermore, we can make the spike rate error arbitrarily small with large values of $T$.

\subsection{Resonate-and-Fire Neurons}

An important alternative building block model to the LIF is the resonate-and-fire (RAF) neuron \cite{izhikevich2001resonate}, also referred to as a ``resonator neuron''. 
Unlike LIFs, which integrate their input stimuli and prefer higher-frequency inputs in order to fire sooner (i.e., a shorter distance between input pulses results in a higher likelihood of an LIF firing), RAFs are driven by the timing of pulses relative to their period of oscillation. 
In other words, if two incoming pulses are near half the period, the input spikes will cancel each other out (and not result in the RAF firing) whereas, if their timing is near one period of oscillation, the spike pulses will add up (and result in the RAF emitting a pulse). For any RAF to fire, it will require stimulation at its resonant frequency (of its subthreshold oscillation). Unlike an LIF, an increase in the frequency of incoming spikes to an RAF might result in a delay of (or even terminate) its own firing.

\subsubsection{Numerical Specification of the RAF}
\label{sec:raf_specification}

The easiest way to think about the dynamics of an RAF neuron is by considering a decaying orbit. As a result, all of the equations used in the description of the RAF neuron's dynamics will be done through the use of complex numbers. To relate a complex number to physical systems, the real component can be thought of as a current-like variable and the imaginary component can be thought of as a voltage-like variable. The state of the RAF neuron, at any point in time, can be represented as follows:
\begin{equation}
    z(t) = x(t) + y(t)i
\end{equation}
and the update (Euler-integrated) dynamics for this neuron is as follows:
\begin{equation}
    \label{eq:raf:base}
    z(t+\Delta t) = (1-b\Delta t + \omega \Delta t i)z(t)
\end{equation}
where $b$ is the rate of attraction towards the origin (or subthreshold dampening factor; this controls how fast the oscillations of the RAF's membrane potential decay back to its resting baseline after a perturbation), and $\omega$ is the frequency of oscillation (this controls the ``bouncing speed'' -- a higher $\omega$ results in more rapid oscillations away from the baseline whereas lower yields a slower, lazier wave); both of these are positive non-zero values.\footnote{In general, an RAF will ``resonate'' (or fire) if incoming pulses hit at the rhythm governed by $\omega$; after a shock by input pulse, $\omega$ defines how many times that the voltage will swing above/below the RAF's baseline value before $b$ works to kill off the movement.} 
The path to obtaining to this equation is as follows: 
\begin{align}
    \nonumber \dfrac{dz}{dt} &= (-b + \omega i) z(t_0) \\
    \nonumber \int^{z(t_0+\Delta t)}_{z(t_0)} d(z) &= (-b + \omega i) z(t_0) \int^{t_0+\Delta t}_{t_0} dt \\
    \nonumber z(t_0 + \Delta t) - z(t_0) &= (-b + \omega i)z(t_0) (t_0 + \Delta t - t_0) \\
    \nonumber z(t_0 + \Delta t) &= (-b + \omega i)z(t_0)(\Delta t) + z(t_0) \\
    \nonumber z(t_0 + \Delta t) &= z(t_0) ((-b + \omega i)(\Delta t) + 1) \\
    z(t_0 + \Delta t) &= z(t_0) (1-b \Delta t + \omega \Delta ti). 
\end{align}
We remark that one can alternatively represent the above RAF dynamics in terms of two non-complex-valued ODEs, essentially constructing a dampened oscillator; however, the single Equation \ref{eq:raf:base} offers a compact to run the simulation as many modern-day linear algebra packages offer native complex (dual-)number support.

\begin{figure}
\centering
\begin{subfigure}{0.495\textwidth}
    \includegraphics[width=\textwidth]{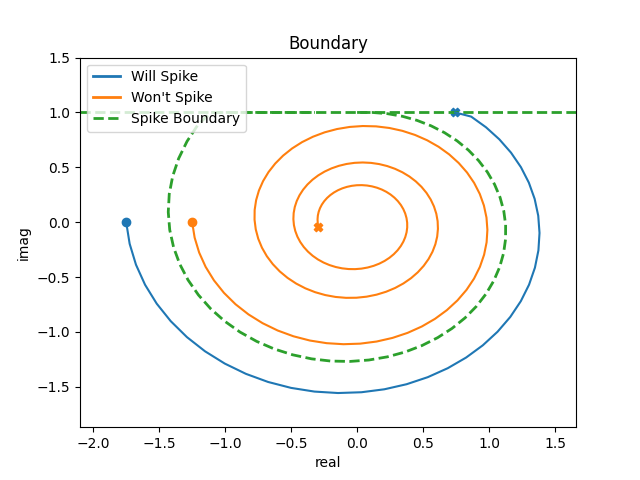}
    \caption{Depicted is a set of ``healthy'' RAF hyper-parameters: $b = 1$, $\omega = 7.5$, and $\Delta t = 0.015$ ms. If the neuron starts outside the spike boundary, it will eventually reach it and spike; whereas, if it starts inside the spike boundary, it will never pass the threshold.}
    \label{fig:models:ref:boundary}
\end{subfigure}
\hfill
\begin{subfigure}{0.495\textwidth}
    \includegraphics[width=\textwidth]{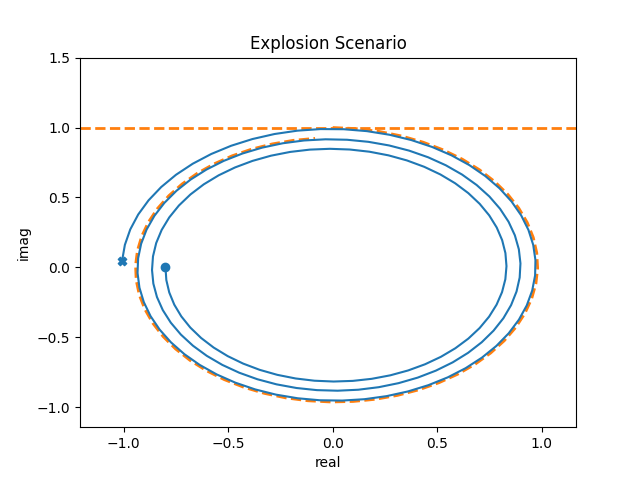}
    \caption{Depicted is a set of ``unhealthy'' RAF hyper-parameters: $b = 0.33$, $\omega = 7.5$, and $\Delta t = 0.015$ ms. Observe that, if thresholding is disabled, the dynamics of the RAF exhibit an increasing magnitude over time (not yielding expected RAF oscillatory dynamics).}
    \label{fig:models:ref:explode}
\end{subfigure}      
\caption{This set of figures shows both healthy (\ref{fig:models:ref:boundary}) and exploding (\ref{fig:models:ref:explode}) with relation to hyperparameter. In both of these figures the neuron dynamics start at the circle point and end at the cross.}
\label{fig:models:raf:dynamics}
\end{figure}

The two key RAF hyper-parameters -- $b$ and $\omega$ -- are both straightforward in how they function yet have to be chosen carefully such that the dynamics of this neuron do not become unstable. Specifically, when this instability occurs, the voltage and current parameters follow exponential growth instead of decay and, as a result, the neuronal model diverges (or ``breaks'') as seen in Figure \ref{fig:models:raf:dynamics}. 
However, there is a mathematical check to detect if this will happen; the check leverages the fact that, after a single time-step, with no input, the magnitude of the vector $z(t+\Delta t)$ must be smaller than the magnitude of $z(t)$. To solve for this useful check, start by using the definition in Equation \ref{eq:raf:base} and then solving the resulting inequality as follows:
\begin{align}
    \label{eq:raf:hyper}
    \nonumber ||z(t+\Delta t)|| &< ||z(t)|| \\
    \nonumber ||z(t) (1-b \Delta t + \omega \Delta ti)|| &< ||z(t)|| \\
    \nonumber ||(1-b \Delta t + \omega \Delta ti)|| &< 1 \\
    \nonumber (1 - b\Delta t)^2 + (\omega \Delta t)^2 &< 1 \\
    \nonumber 1 - 2b\Delta t + b^2\Delta t^2 + \omega^2 \Delta t^2 &< 1 \\
    \nonumber b^2\Delta t^2 + \omega^2 \Delta t^2 &< 2b\Delta t \\
    \frac{b^2 + \omega^2}{b} &< \frac{2}{\Delta t}.
\end{align}
As a result, one would choose values for $b$ and $\omega$ such that the above inequality is satisfied under a given experimental integration time constant $\Delta t$.

\begin{figure}
    \centering
    \includegraphics[width=0.615\linewidth]{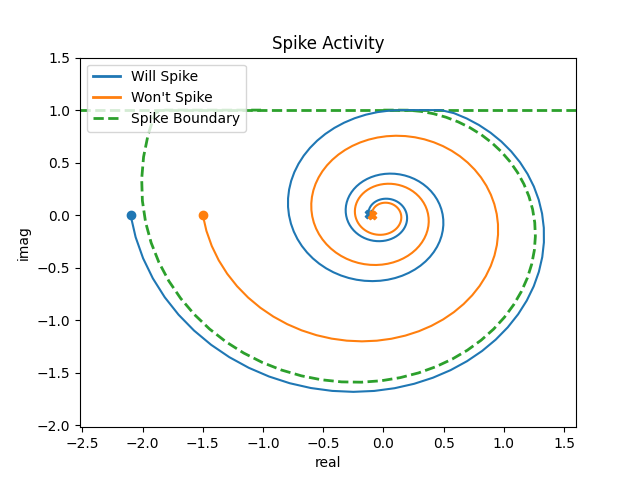}
    \caption{Here, we visually depict the process of repolarization in the RAF unit. It can be seen that, when the neuron passes its membrane threshold, it is held there until the real number component passes into the negative range and, as a consequence of this, the neuron falls below the spike boundary / voltage threshold.}
    \label{fig:models:raf:spike}
\end{figure}

Given both the RAF definition (\ref{eq:raf:base}) and the hyper-parameter bound (Equation \ref{eq:raf:hyper}), all the requisite tools required to simulate the dynamics of a RAF neuron are available. 
There is only one more step to take: handling when the RAF unit is to produce a spike pulse. Notably, the RAF neuron does have a threshold similar to that of an LIF neuron; in contrast, this threshold is defined as a positive value along the imaginary number axis. As a result, the equation used to govern if there is a spike is as follows: 
\begin{equation}
    S(t) = \begin{cases}
        1 & \Im(p(t)) > \theta(t) \\
        0 & \text{otherwise}
    \end{cases}. 
\end{equation}
Another interesting behavioral element of the RAF that is important to consider is how it repolarizes. Specifically, unlike the LIF neuron, which resets the voltage to a specific reset value after spike-emission (i.e., immediate or instantaneous repolarization), in the RAF model, the imaginary part of the neuron is clamped to the threshold and held there until the real part passes into the negative range. 
See Figure \ref{fig:models:raf:spike} for a visual representation of this particular behavior/process. 

\begin{figure}
    \centering
    \includegraphics[width=0.615\linewidth]{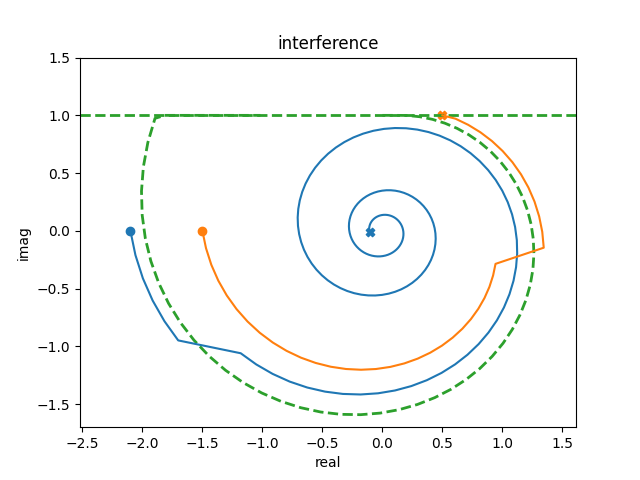}
    \caption{In this diagram, we visualize both constructive and destructive interference in RAF neurons. One of the neurons would spike if no input was provided; however, input was provided at the ``wrong'' time and, as a result, the neuron ends up stopping a spike. The other neuron, in contrast, gets pushed past its own membrane threshold under the same sized input current, resulting in a spike.}
    \label{fig:models:raf:interference}
\end{figure}

The final part of the RAF dynamics that has not been covered yet is the injection of electrical current into the neuron itself. To do this in practice, the real component $x(t)$ of the RAF neuron is directly incremented by the input current $j(t)$, as seen in Figure \ref{fig:models:raf:interference}. This means that, depending on where the neuron is in its rotation cycle, this addition to the real part will either be constructive or destructive to the RAF's overall dynamics.

\paragraph{Biological Starting Point:} 
In terms of neurobiological experimental studies, there is far less available to guide one in how the constants that govern the RAF should be set. 
Nevertheless, in practice, many of the shared core constants that the RAF shares with the LIF can be set much like one would for the LIF. For instance, the voltage reset $v_{\text{reset}}$ (for hyperpolarization) and resting membrane potential $v_{\text{rest}}$ (for repolarization) can be set to a range of values, such as $[-70,-65]$ mV whereas the voltage threshold base value $\theta_{\text{base}}$ can also be set to around $-55$ mV. 
The value for the resonant angular (eigen)frequency $\omega$ characteristic of the RAF is set according to the type of neuron's subthreshold membrane (potential) oscillations that one wants to model (which vary widely in the brain). If one wishes to model neurons that exhibit lower-frequency oscillations, such as those that occur in the inferior olive and the hippocampus, then values around $\omega = [5, 10]$ Hz are reasonable; for neurons that exhibit higher-frequency oscillations, e.g., cortical neurons, then higher values will be required. 
The other meta-parameters of the RAF tend to be dimensionless variables  mostly to control the dampened oscillatory behavior of the underlying two-dimensional system, and thus are not set in accordance to biological experimental data or observations. 

\subsubsection{Tuning RAF Dynamics}
\label{sec:tuning_rafs}

\begin{figure}
\centering
\begin{subfigure}{0.495\textwidth}
    \includegraphics[width=\textwidth]{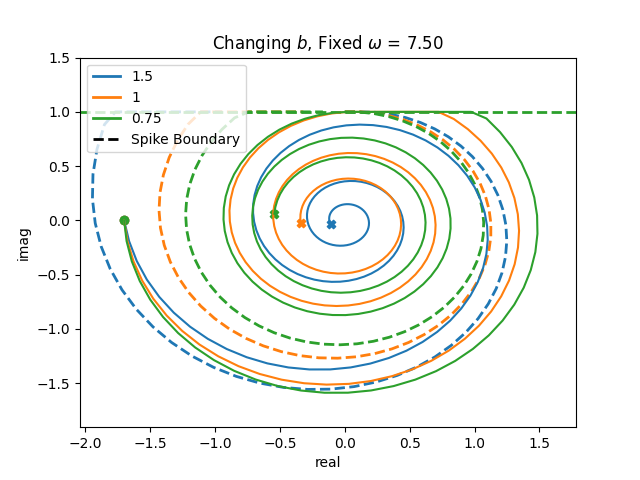}
    \caption{
    \textbf{The effect of changing the hyperparameter $b$ under fixed $\omega$ for RAF neuronal units.} Observe that, with the larger the value of $b$, more current (and, consequently, membrane voltage) will be required for an RAF neuron to be inside the range needed to emit a spike pulse.}
    \label{fig:models:ref:tb}
\end{subfigure}
\hfill
\begin{subfigure}{0.495\textwidth}
    \includegraphics[width=\textwidth]{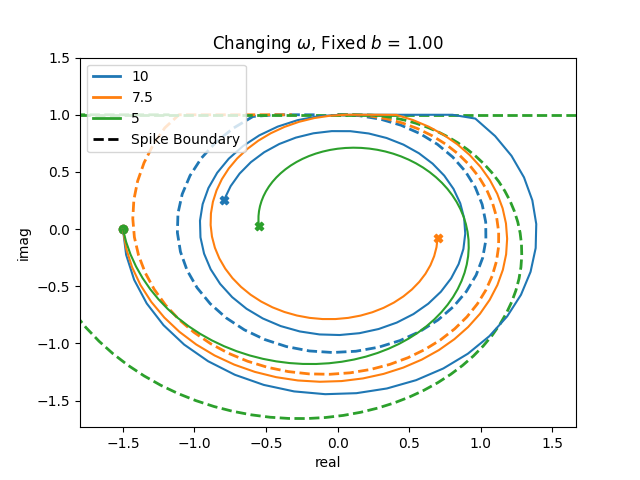}
    \caption{
    \textbf{The effect of changing the hyperparameter $\omega$ under fixed $b$ for RAF units.} Larger values of $\omega$ will pull the voltage threshold needed from injected current to produce a spike. Observe that the number of revolutions completed by the larger $\omega$ is slightly further than those with a smaller $\omega$. }
    \label{fig:models:ref:tw}
\end{subfigure}        
\caption{This set of figures illustrate the effects of changing the RAF unit's two different hyper-parameters ($\omega$ and $b$) and their impact on the dynamics of a group of three neurons. In both figures, all neurons are run from the same initial conditions and for the same number of time steps. It is important to note that the integration time constant $\Delta t$ for these trials was set to $0.015$ ms. This means that the only variations across the three neurons in each figure is the designated hyper-parameter setting.}
\label{fig:models:raf:hyperparams}
\end{figure}

When tuning a RAF neuron, there is not a great deal that one needs to do when compared to an LIF. 
Since there are only two key hyper-parameters, and they are loosely dependent on one another, the turning process can become a rather straightforward process (provided that good values are chosen to avoid instability, as discussed earlier). The first hyper-parameter to examine is $b$, or the rate of attraction towards rest. 
The larger that this value is set to, the faster that the neuron will be pulled towards the center; as a byproduct, if there is a large degree of sparsity in the (sensory) inputs presented to the neuron, a lower value would be expected/useful since, otherwise, the neuron will return to rest in between input spikes. 
The second hyper-parameter to tune is $\omega$ -- this value, as discussed before, represents the (eigen)frequency of the neuron. 
There are two ways that this is tuned; first, if the model is operating on the same frequency across all the neurons (which is a common pattern), then this value will just end up being fixed and, as a result, no tuning is needed. However, if there are experimental scenarios where this is not the case, some degree of tuning will be required. The main impact that this value has on the RAF's dynamics is that it affects both the needed input frequency range as well as the potential frequency of spikes coming out of the neuron. As such, this value is often turned in conjunction with respect to an entire (SNN) model as opposed the to individual single neuron in isolation. The effects of these two hyperparameters can be seen in Figure \ref{fig:models:raf:hyperparams}

\subsection{The Spike Response Model}
\label{sec:spike_response_model}

Many of the spiking neuron models discussed above are described in the form of differential equations whose solutions require numerical approximations. As a result, the simulation of these types of models are subject to errors introduced by time discretization (see Section \ref{sec:time_discretization_error}).  Moreover, simulated-time, loop-based (clock-driven) simulation of such models can be very slow when the time step is small and can also be incredibly inefficient when SNN activity is sparse. 

Event-driven simulation offers an alternative that notably only requires SNN state equations to be updated when neurons spike \cite{mo2021edha,mo2022evtsnn}. One type of spiking neuron model that is particularly well-suited for event-driven simulation is the spike response model (SRM), introduced by Gerstner \cite{gerstner1995time}. In fact, one can think of the SRM more as a framework or family of models that explicitly describes the impact of pre-synaptic spikes on a post-synaptic neuron's membrane potential:
\begin{equation}
    v_{i}(t)=\eta(t-t_{i,\hat{k}})+\sum\limits_{j}w_{i,j}\sum\limits_{k}\epsilon_{i,j}(t-t_{i,\hat{k}},t-t_{j,k})+\int\limits_{0}^{\infty}\kappa(t-t_{i,\hat{k}},t')I^{ext}(t-t')\mathrm{d}t'
\label{eqn:srm}
\end{equation}
where $\eta$, $\epsilon$, and $\kappa$ are kernels that describe the spike shape, post-synaptic potential response, and response to external current, respectively. The index $i$ refers to the post-synaptic neuron and the indices $j$ refer to the pre-synaptic neurons.  The term $t_{i,\hat{k}}$ is the last time that the post-synaptic neuron spiked, whereas $t_{j,k}$ is the $k^{\mathrm{th}}$ time that the pre-synaptic neuron $j$ spiked. 
Finally, the value $w_{i,j}$ refers to the synaptic weight between neuron $j$ and neuron $i$. In practice, every time that a pre-synaptic neuron fires, Equation \ref{eqn:srm} can be evaluated for neuron $i$ to find the new membrane potential.

If the network activity is sparse, then the number of evaluations will be much smaller than that of a time-based simulation. Note, also, that when $v_{i}(t)$ reaches a threshold $\theta(t,t_{i,\hat{k}})$, then the post-synaptic neuron will emit a pulse. 
We may take advantage of this fact and reformulate the LIF voltage (Equation \ref{eqn:lif_voltage}) such that we may derive the following LIF-SRM (see supplementary material for the full derivation): 
\begin{align}
    \tau_m \frac{dv_i(t)}{dt} = \eta(t - t_{i,\hat{k}}) + \sum_j w_{i,j} \sum_k \epsilon(t - t_{i,\hat{k}},\, t - t_{j,k}) + \int_0^{\infty} \kappa(t - t_{i,\hat{k}},\, t')\,I^{ext}(t - t')\,dt' . \label{eqn:derived_srm}
\end{align}
Note that the SRM constructed in Equation \ref{eqn:derived_srm}, to be used to instantiate an LIF, indicates that it entails evaluation of an integral, which entails in implementation maintaining the history of input event spike times. 

\begin{figure}[!t]
\centering
\begin{subfigure}{1.0\textwidth}
\centering
    \includegraphics[width=0.8\textwidth]{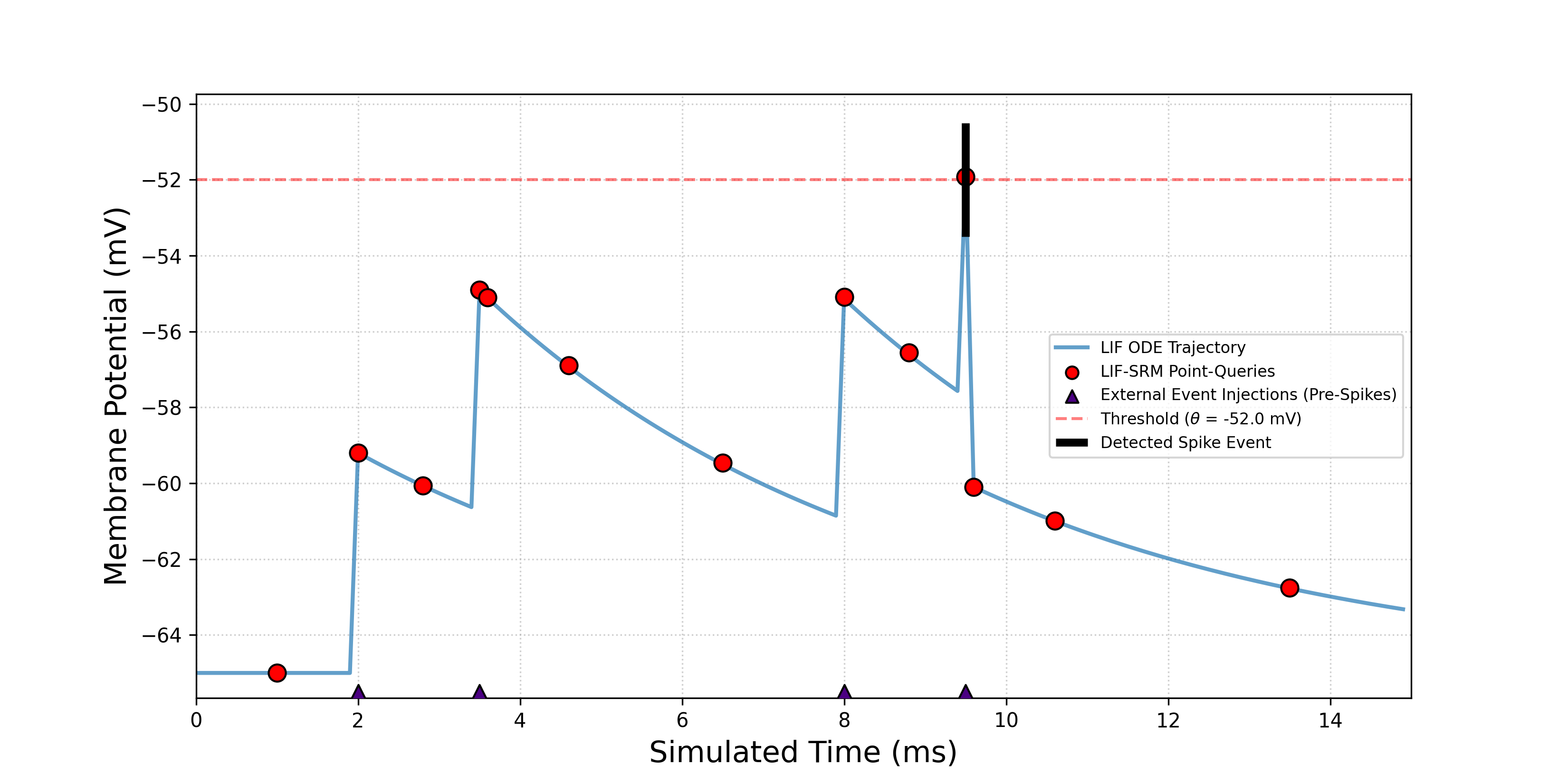}
    \caption{LIF ODE vs. LIF-SRM Sampled Points.}
    \label{fig:models:ref:srm:lif}
\end{subfigure}
\hfill
\begin{subfigure}{1.0\textwidth}
\centering
    \includegraphics[width=0.8\textwidth]{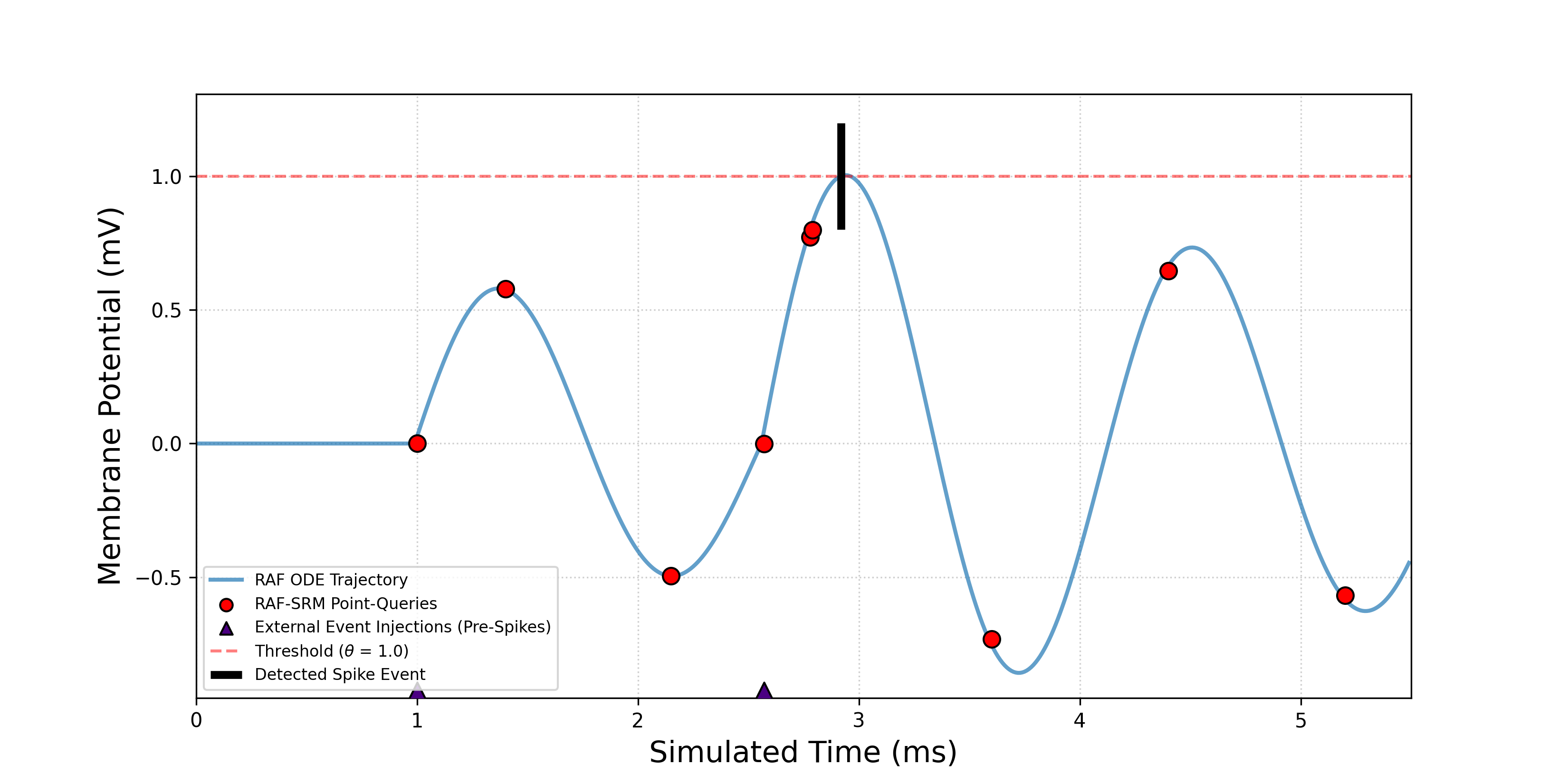}
    \caption{RAF ODE vs. RAF-SRM Sampled Points.}
    \label{fig:models:ref:srm:raf}
\end{subfigure}        
\caption{
Here, we visualize sampled queries of: 
\textbf{(a)} the leaky integrate-and-fire spike-response model (LIF-SRM) overlaid against the continuous trajectory induced by a comparable LIF ordinary differential equation (LIF ODE); and,
\textbf{(b)} the resonate-and-fire SRM (RAF-SRM) overlaid against the continuous trajectory of its comparable RAF ODE (RAF ODE). Note that, over a window of time where several pre-synaptic input events are injected at particular moments, the SRM formulations of the LIF and RAF facilitate a sparse, event-based form of calculation; in other words, they facilitate an efficient means of querying at a moment in time obtains an analytical voltage value. 
}
\label{fig:models:srm_dynamics}
\end{figure}

For the RAF, an SRM-formulation follows a similar format as the LIF; however, the choice of the external input kernel $\kappa(t - t_{i,\hat{k}},\, t')$, the synaptic input kernel $\epsilon(t - t_{i,\hat{k}},\, t - t_{j,k})$, and the spike-reset kernel $\eta(t - t_{i,\hat{k}})$ must be made in accordance to obtain RAF-like dynamics \cite{aoun2022resonant}. This involves designing damped sinusoidal kernels for both $\kappa(t - t_{i,\hat{k}},\, t')$ and $\eta(t - t_{i,\hat{k}})$ (such as  oscillatory, decaying alpha-like functions, instead of decaying exponential functions used for the LIF); this yields what is known as the cumulative SRM \cite{richardson2003subthreshold,aoun2022resonant}. 
For instance, building the RAF within Equation \ref{eqn:derived_srm} involves making kernel design choices  such as:
\begin{align}
\eta(t - t_{i,\hat{k}}) &= \text{H}(t - t_{j,k}) \text{H}(t_{j,k} - t_{j,\hat{k}}) \exp\big(-b (t - t_{j,k}) \big) \sin(\omega (t - t_{j,k}) ) \notag \\ 
\kappa(t - t_{i,\hat{k}}, t^\prime) &= \text{H}(t^\prime) \text{H}(t - t^\prime - t_{i,\hat{k}}) \exp(-b t^\prime) \cos(\omega t^\prime) \notag \\ 
\eta(t - t_{i,\hat{k}}) &= \text{H}(t - t_{i,\hat{k}}) v_{\text{reset}} \exp(-b (t - t_{i,\hat{k}}) \cos(\omega (t - t_{i,\hat{k}}) \notag 
\end{align}
where $\omega$ is the angular frequency and $b$ is the dampening factor (as in the RAF; Equation \ref{eq:raf:base}) and $\text{H}(\cdot)$ is the Heaviside function (used to ensure that our SRM only counts pre-synaptic pulses that occur after the last post-synaptic reset). 

Note that, in both cases, the SRM definition expresses the internal states of the neuronal units in terms of complex, memory-intensive integrals (i.e., we must maintain a history of input spike times). Nevertheless, since both of these particular kernels adhere to exponential decay (the LIF) or dampened harmonic oscillation (the RAF) dynamics, these integrals can be collapsed into closed-form 
transformations. Specifically, we may track particular variables (such as the input current administerd to the LIF; or the current that drives the voltage and angular momentum variables of the RAF) via traces in tandem with recent spike-times to design full event-based kernel models while sidestepping the overhead cost of inefficient historical tracking of all spike-times. In Figure \ref{fig:models:srm_dynamics}, we visualize the operation of SRMs formulated for both LIF and RAF dynamics, highlighting the SRMs' event-based nature; note that an SRM computes the state of its dynamics at a queried point as opposed to being run continuously as in the case of a spiking model in its ODE format.

\section{On Inhibition and Excitation in Spiking Networks}
\subsection{Biological Framing}

To build effective and useful SNN circuit models, it is important to consider certain details related to the overall structure and organization of their internal neuronal units and synaptic connections. One such detail is the type of structures that give rise to important effects such as inhibition, excitation, and the relationship between these two \cite{zhou2018synaptic}. Excitatory and inhibitory pressures often entail a balancing act between signals that stimulate (or increase/elevate) the activity of neuronal cells and signals that depress (or decrease/dampen) the activity of neuronal cells. 

Excitation is generally a key driver of the transmission of activity across spiking neurons; in SNNs, synaptic projections are introduced to ferry signals across connected cells; pulses that leave the axons of cells enter into the dendrites of others and these signals are accumulated/summed, representing the positive/non-negative signals known as \emph{excitatory post-synaptic potentials} (EPSPs) that will make post-synaptic neurons more likely to produce an action potential. In addition to the forward-projective form of EPSPs, in SNNs, we also consider another form known as self-excitation; in this case, we model the fact that when a spiking neuron emits an action potential, it potentially increases the chance that it will fire again in the near future (this is often modeled via a self-recurrent loop).

With respect to inhibitory pressure, or the accumulation of the negative signals known as \emph{inhibitory post-synaptic potentials} (IPSPs; these make a post-synaptic neuron less likely to emit an action potential), a particular form that is often emulated is lateral inhibition. This is a structural format that allows for an excited neuronal cell to reduce the activity of others that are located nearby, i.e., its laterally-connected neighbors. 
Biologically, cells that make use of this type of inhibition the most are found in the thalamus or cerebral cortex, often leading to the construction of lateral inhibitory networks. For instance, in the context of perception \cite{von2017sensory} in mammalian retina, lateral inhibition is an important means of allowing the brain to sharpen and perceive contrast within visual stimuli. In a dark environment, when a small bit of light enters the eye, photo-receptive rods in the center of the stimulus will process and transmit ``light'' pulses while other rods outside of the stimulus will transmit ``dark'' pulses as a result of the (lateral) inhibitory signals produced by retinal horizontal cells; 
this creates a contrast between the light and dark aspects of perceived input.

When taken together, both EPSPs and IPSPs facilitate the modeling of the balance of excitation and inhibition \cite{zhou2018synaptic}, i.e., E/I balance, or the relative contributions of excitatory and inhibitory synaptic inputs which trigger pulse emission in neurons. Note that ``balance'' is achieved if, under a range of specified conditions, the ratio between both excitation and inhibition is constant. Biologically, a balancing, regulatory effect in brain structures such as the (neo)cortex is achieved via interneurons, which are responsible for producing inhibitory pressures. However, the physical quantity of the types of (inhibitory) neurons is a small fraction of the total population; for example, in the neocortex, a commonly observed ratio excitatory-to-inhibitory neurons is approximately $80$\%-$20$\% \cite{noback2005human}. Generally, these observations in neurobiology translate into some form of constraint or model-level design, the basic forms of which this section will study.

\subsection{Constructing the Spiking Network Model}


Going from atomic elements, e.g., encoding unit, spike cell, to a full SNN model entails consideration of the (virtual) topology that one wants to construct. This is the consideration of the ``assembly'', or overall organization of units into groups, layers, or populations. This is the step where the modeler constructs the architecture of the SNN; for instance, designing a model with multiple populations of LIFs organized in a hierarchical or heterarchical fashion. In this section, we will consider one of the simpler possible formats -- a two-layer SNN, where a sensory input/encoding layer is connected to a hidden layer of spiking neuronal units via one set of synaptic projections (i.e., a single synaptic weight matrix). The sensory input units will generate EPSPs that will drive the hidden LIFs units; notice that this model will only include sensory-to-hidden excitatory projections and not sensory-to-hidden inhibitory ones (though the modeler might consider the integration of these to generate IPSPs from the input, depending on the type of neural circuit to be modeled). 

The above design choice means we are modeling the electrical current to be injected into the hidden layer -- $\mathbf{i}^1(t)$, where the superscript $1$ denotes the first hidden layer index (where any layer of a hierarchical structure is indexed via $\ell$; the sensory input layer is $\ell=0$) -- as a function of the projection of the sensory layer's spikes. This would yield the following model $\mathbf{i}^1(t) = \mathbf{W}^1 \cdot \mathbf{s}^0$ in terms of a full matrix of synaptic efficacies or, in terms of the $i$-th neuron in layer $1$ (in order to better align with the single-neuron specification of Section \ref{sec:lif_specification}), we have the following: 
\begin{align}
    i^1_i(t) = \sum^{N_0}_{j=1} W^1_{ij} s^0_j = \sum_{j: s^0_j = 1} W^1_{ij}
\end{align}
where $N_0$ is the number of neurons in the sensory $\ell=0$ layer. Note that, in the above, we present in the rightmost side of the equality in terms of only a summation of particular values within matrix $\mathbf{W}^1$ given that spikes emitted from layer $\ell=0$ are binary; this allows us to side-step using the multiplication operation and further only consider the non-zero values in $\mathbf{s}^0$ as part of the computation of injected input current into neuron $i$ of hidden layer $\ell=1$. This form of computing electrical current also highlights the lower (power) cost of an SNN, since a dense set of multiplications is no longer required (i.e., multiplying two numbers generally consumes significantly more energy than adding them). 

The last initial consideration the modeler will want to make is how to initialize the values of the synaptic efficacies within matrix $\mathbf{W}^1$. In practice, a reasonable first-choice is to initialize the values from a non-negative, sparse uniform distribution, i.e., $W^1_{ij} \sim \mathcal{U}(a, b)$ where $a \geq 0$ and $b > a$. It is preferable to use a uniform distribution over a Gaussian distribution (centered around a high enough non-zero value) as a useful way to create good initial dynamics of a population of neurons (in the hidden layer) is to ensure that the variance of their activities is maximized, i.e., we want high variance in the neural activity values in the population to promote higher diversity and a better usage of the dimensional space of the hidden (sparse code) space. 
This is also related, in some sense, to the effective dimension as often closely examined in encoder-only models in self-supervised learning \cite{geiping2023cookbook}; we ultimately want a high effective dimension, meaning that a good proportion of the vector space induced by the hidden layer is utilized or at least plays some role in the SNN's evolution over time. 

A final important detail is to decide on the degree of sparsity in $\mathbf{W}^1$; this can be done by sampling a binary mask $\mathbf{M}^1$  of the same shape as $\mathbf{W}^1$ and masking the forward projection matrix -- $\mathbf{W}^1 \leftarrow \mathbf{W}^1 \odot \mathbf{M}^1$. 
Note that each element of $\mathbf{M}^1$ is sampled from a Bernoulli distribution, under a probability $p_m$, i.e., $M^1_{ij} \sim \mathcal{B}(M^1_{ij}=1; p)$. This initialization scheme provides control to the modeler to determine the degree of sparsity (via $p_m$) desired in the structure of the EPSPs emitted from layer $\ell=0$ to $\ell=1$.

\subsection{Inhibitory-Excitatory Design Patterns}

To create effective SNN systems, as motivated above, one of the most important effects to model is the balance of excitation and inhibition \cite{zhou2018synaptic}. 
In this section, we describe and conduct several experiments that study the ratio of excitatory-to-inhibitory neurons; concretely, we examine the biological ratio of $80$\%-$20$\% (as known in the neocortex \cite{noback2005human}) as well as equally-sized populations of both. Furthermore, we experiment with types of random synaptic connectivity that connect the groups of excitatory and inhibitory neurons in our LIF-centric SNN models.

\subsubsection{Effects of Equally-sized Populations of Inhibitory and Excitatory Neurons}
\label{sec:equal_ei_populations}

In computational neuroscience models, one common form of inhibition is a fully-connected layer of excitatory neurons that is laterally wired to the same number of neurons in a separate inhibitory layer; this is a biophysical design pattern used in well-known models such as the SNN of \cite{diehl2015unsupervised}. The matrix of cross-layer connections between these two layers of neurons (in either direction, i.e., excitatory-to-inhibitory and inhibitory-to-excitatory) normally takes on the form of either a hollow matrix (one-to-many connectivity) or of an identity matrix (one-to-one connectivity). 
The connectivity matrix between the layers will always consist of one of each of these types of matrices, with each setup having different implications on the dynamics that govern the paired excitatory-inhibitory groups.

\begin{figure*}[!t]
    \centering
    \begin{subfigure}[b]{0.45\textwidth}
        \includegraphics[width=1\textwidth]{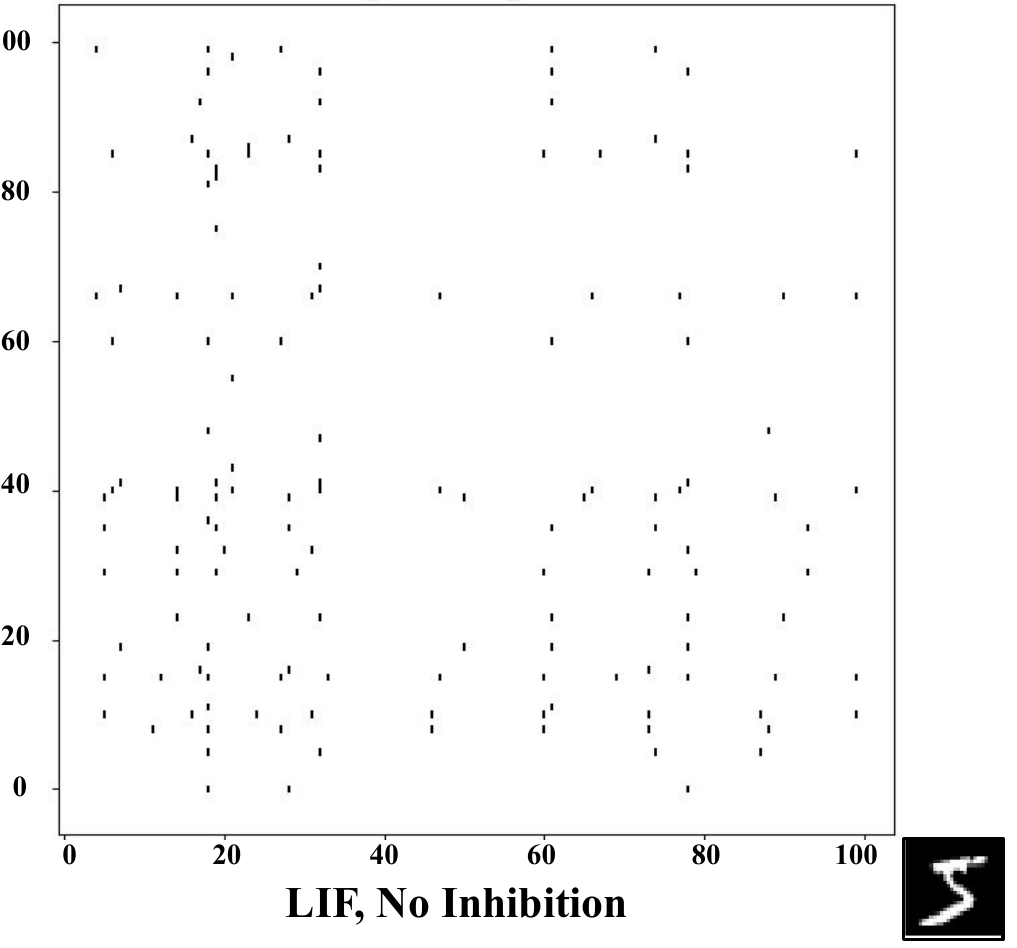} 
        \caption{}
        \label{fig:lif_no_inh}
    \end{subfigure}
    \begin{subfigure}[b]{0.45\textwidth}
        \includegraphics[width=1\textwidth]{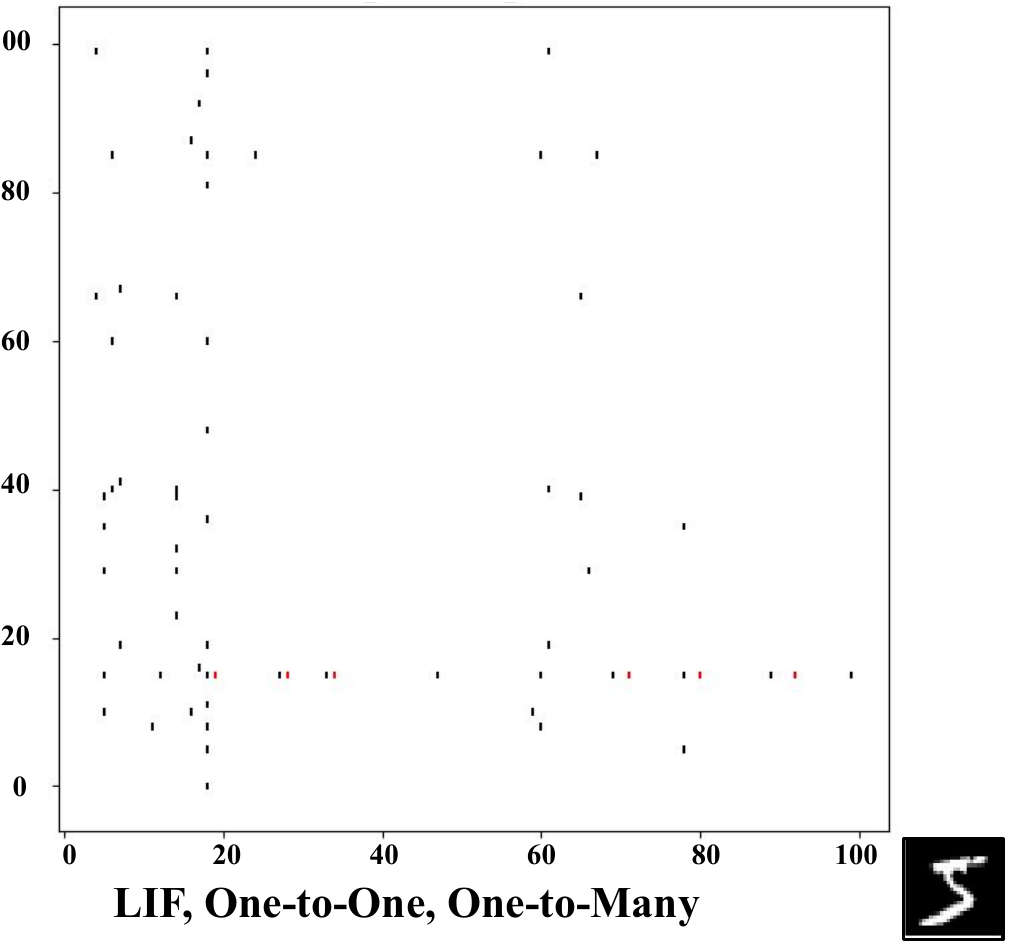} 
        \caption{}
        \label{fig:lif_ooom_result}
    \end{subfigure}\\
    \begin{subfigure}[b]{0.45\textwidth}
        \includegraphics[width=1\textwidth]{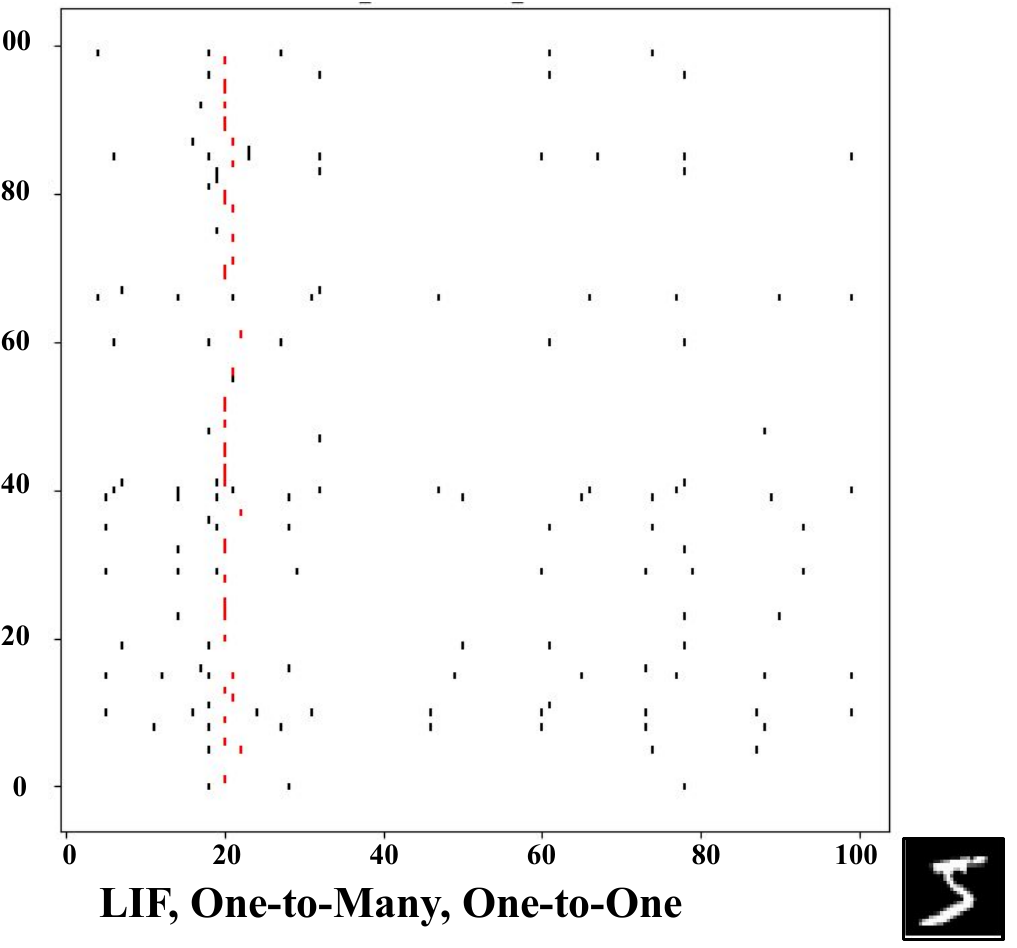} 
        \caption{}
        \label{fig:lif_omoo_result}
    \end{subfigure}
    \begin{subfigure}[b]{0.45\textwidth}
        \includegraphics[width=1\textwidth]{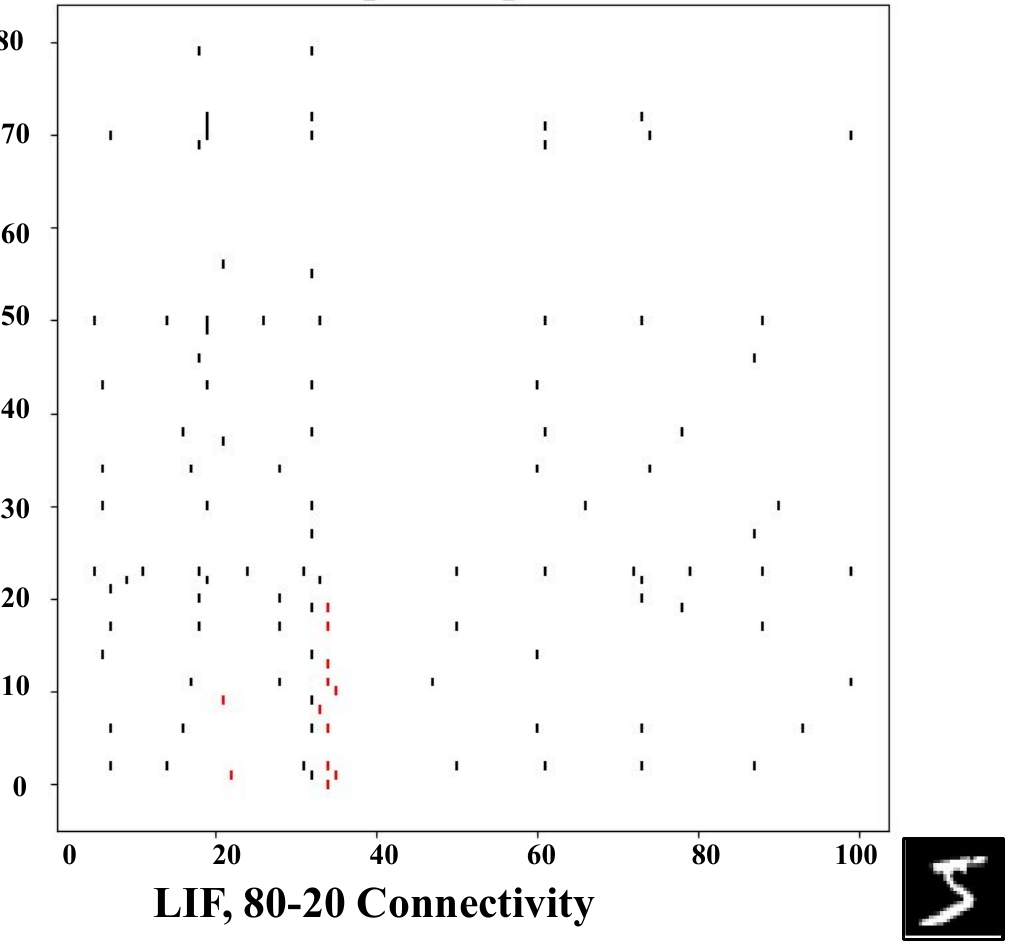} 
        \caption{}
        \label{fig:lif_80_20_result}
    \end{subfigure}
    \caption{The effect that lateral synaptic connectivity -- one-to-one/one-to-many, one-to-many/one-to-one, and sparse $80$:$20$ split connectivity (with $80$:$20$ ratio of excitatory-to-inhibitory units) -- has on dynamics of two populations (one excitatory and one inhibitory) of LIFs. In all plots, black ticks correspond to events in excitatory neurons; likewise, red ticks correspond to inhibitory neuron events. For the $80$:$20$ split, there will only be red ticks from neuron index $0$ to $20$ (for inhibitory neurons).}
    \label{fig:results_lateral_connectivity}
\end{figure*}

\paragraph{One-to-One, One-to-Many Lateral Connectivity} 
For this trial an SNN model with a one-to-one set of connections (a diagonal matrix) from the excitatory neurons to the inhibitory neurons, followed by a one-to-many set of connections (a hollow matrix) from the inhibitory neurons to the excitatory neurons, was created where all of the neurons are LIFs. The encoding sensory input layer was made up of Poisson neurons all with a target frequency of $63.75$.

When using LIF neurons the effect that this inhibition pattern produces is similar to a winner-take-all (WTA) scheme, i.e., only one or a few neurons fire over a large period time. In practice, this effect is achieved through the quick firing of a few excitatory neurons which causes their corresponding inhibitory neurons to also fire quickly. This quick firing of a select group of inhibitory neurons dampens the stimulation of the other excitatory neurons. Since the spiking excitatory neurons that initially spike will receive less of a dampening force (as compared to those that did not spike initially), they will, in turn, fire faster again, ultimately producing the WTA-like output. This result can be observed in Figure \ref{fig:results_lateral_connectivity}, specifically when comparing the model with no inhibition in Figure \ref{fig:lif_no_inh} to the model with inhibition in Figure \ref{fig:lif_ooom_result}.

When tuning this particular structure, there are several factors to note: 
\textbf{1)} since there is significantly less stimulus being produced by the excitatory neurons, the resistance values ($R$) and the time constant ($\tau$) of the inhibitory structure need to be increased relative to the excitatory structure in order to produce any spiking activity at all; 
\textbf{2)} adjusting the resistance of the excitatory-to-inhibitory connections will decrease the amount of excitatory activity needed to produce inhibitory activity; and, finally,   
\textbf{3)} adjusting the resistance of the inhibitory-to-excitatory connections increases the inhibiting effect of the structure and pushes it much closer to WTA behavior.

\paragraph{One-to-Many, One-to-One} 
The alternative way of setting up the SNN model, given an equal amount of excitatory and inhibitory neurons, is with a one-to-many set of synaptic connections (a hollow matrix) from the excitatory neurons to the inhibitory neurons and then a one-to-one set of connections (a diagonal matrix) from the inhibitory neurons to the excitatory neurons. When this model is built using LIF neurons, the inhibition pattern that is produced is akin to denoising. Since the inhibitory neuron that is used to inhibit/suppress a specific neuron in the excitatory layer is stimulated from all of the other neurons, if the excitatory neuron being inhibited is on the edge of firing (or firing very slowly, without inhibition) it will be quieted; as such, it will never spike. 
The de-noising effect can be observed in the LIF neurons by comparing Figure \ref{fig:lif_no_inh} and Figure \ref{fig:lif_omoo_result}.

\subsubsection{The 80-20 split}
\label{sec:80_20_ratio}

The construction of the ratio of $80$\% excitatory neurons to $20$\% inhibitory neurons was not chosen at random but instead motivated by neurophysiological and computational studies of the proportion of excitatory to inhibitory neuronal cells in particular brain regions/structures (e.g., the neocortex \cite{noback2005human,deco2014local,alreja2022constrained})\footnote{The more general principle is that there are more excitatory neurons in relation to inhibitory ones; other ratios that have also been examined include $3$:$1$ and $9$:$1$ across animal species.}. It is notable that in an adult mouse brain, the excitatory-to-inhibitory ratio is approximately $8$:$2$ \cite{rodarie2022method}. 

Under this setup, there cannot be a one-to-one set of synapses connecting together the neurons of this pair of populations as the connection matrices are always densely connected with a roughly $20$\% dropout rate (i.e., about $20$\% of values are set to zero). The effects of this structure of neurons are also not mirrored in either of the two previous structures. Instead, an $80$\%-$20$\% format effectively functions as a threshold applied to the total activity from excitatory neurons before inhibiting all of them; this effectively slows down or stops future neural activity. This dampening leads to an initial spike pattern found in the excitatory neurons to carry more weight as, after a short amount of time, the spiking decreases significantly. See Figure \ref{fig:lif_80_20_result} for a raster plot depicting the result of our experimental study of the effect that an $8$:$2$ ratio wiring pattern has on dynamics. 

\subsection{General Guidelines on Lateral Connectivity}

In general, when constructing inhibition for SNNs, the amount of stimulus to transfer (across layers) as well as the intended use (e.g., task relevance, what the modeler wants to probe/measure) of a layer or population needs to be taken into account. If the population of excitatory neurons are part of a series of layers where they are meant to facilitate (unsupervised) clustering, as well as clean signals, the experimentalist should consider either of the setups that result in a denoising effect, i.e., the one-to-many, many-to-one or the 80-20 split. 
These allow for the highest degree of stimulus to be transferred deeper (more downstream) into the SNN model while still providing the intended effect. 
The many-to-one, one-to-many setup should be used when winner-take-all behavior is desired. Oftentimes, this is quite valuable for a final (or intermediate) output layer of an SNN where the representation produced needs to be correlated with a label or other one-hot encoded signal.

\section{Conclusion}
In this work, we studied and experimentally analyzed practical elements in tuning various forms of input encoding for spiking neural networks, two fundamental building block neuronal cells -- the leaky integrate-and-fire and the resonate-and-fire cell -- used to construct full spiking circuit models, and the effects of various design patterns for assemblies of leaky integrators, particularly with respect to laterally-wired populations of excitatory and inhibitory neuronal cells. 

This empirical, simulation-based work is meant to provide a detailed starting point for researchers, engineers, and practitioners in brain-inspired computing and computational neuroscience who seek to construct their own biophysical neuronal models, particularly those who intend to work with spike-emitting / pulse-based computational units. 
An important future extension of this study should include an examination of the effect that spike-timing-dependent plasticity \cite{bi1998synaptic,dan2004spike} (a very common synaptic plasticity scheme for incorporating learning into an SNN) has on the neuronal dynamics of leaky integrators and resonators, once an SNN circuit has been set up according to the baseline prescriptions, hyper-parameter considerations, and lateral excitatory-inhibitory connectivity patterns studied in this work for a randomly initialized, static model configuration.

\subsection*{Acknowledgments}
Any opinions, findings and conclusions, or recommendations expressed in this material are those of the authors, and do not necessarily reflect the views of the US Government, the Department of War, or the Air Force Research Lab.
\noindent
Approved for Public Release; Distribution Unlimited: PA Case \#:
AFRL-2026-3409

\bibliographystyle{agsm}
\bibliography{ref}

\newpage
\section*{Supplementary Material}

Here, we present our full derivation of the SRM kernel-based equation, yielding the modeling framework introduced in
\cite{gerstner1995time}. 
The general form of this model family directly, explicitly describes the impact of pre-synaptic spikes on a post-synaptic neuron's membrane potential:
\begin{equation}
    v_{i}(t)=\eta(t-t_{i,\hat{k}})+\sum\limits_{j}w_{i,j}\sum\limits_{k}\epsilon_{i,j}(t-t_{i,\hat{k}},t-t_{j,k})+\int\limits_{0}^{\infty}\kappa(t-t_{i,\hat{k}},t')I^{ext}(t-t')\mathrm{d}t'
\label{eqn:srm_appendix}
\end{equation}
where, again, $\eta$, $\epsilon$, and $\kappa$ are kernels that describe the spike shape, post-synaptic potential response, and response to external current, respectively. Index $i$ refers to the post-synaptic neuron and index $j$ refer to the pre-synaptic neurons. The term $t_{i,\hat{k}}$ is the last time that the post-synaptic neuron spiked, whereas $t_{j,k}$ is the $k^{\mathrm{th}}$ time that the pre-synaptic neuron $j$ spiked. 
Finally, the value $w_{i,j}$ refers to the synaptic weight between neuron $j$ and neuron $i$. In practice, every time that a pre-synaptic neuron fires, Equation \ref{eqn:srm_appendix} can be evaluated for neuron $i$ to find the new membrane potential. 

If the network activity is sparse, then the number of evaluations will be much smaller than that of a time-based simulation. Note, also, that when $v_{i}(t)$ reaches a threshold $\theta(t,t_{i,\hat{k}})$, then the post-synaptic neuron will emit a pulse. In order to get the LIF model into the SRM format, we start with the following (a generalization of Equation \ref{eqn:lif_voltage}): 
\begin{equation}
\tau_{m}\frac{\mathrm{d}v(t)}{\mathrm{d}t}=-v(t)+R\sum\limits_{j}w_{i,j}\sum\limits_{k}\alpha(t-t_{j,k})+RI^{ext}(t)=-v(t)+f(t)
\end{equation}
and solve it in the following manner:
\begin{align}
\nonumber \tau_m \frac{dv}{dt} &= -v + f(t), \qquad v(t=t_0) = a \\[6pt]
\nonumber \frac{dv}{dt} + \frac{v}{\tau_m} &= \frac{f(t)}{\tau_m} \\[6pt]
\nonumber e^{t/\tau_m}\left[\frac{dv}{dt} + \frac{v}{\tau_m}\right] &= \frac{e^{t/\tau_m} f(t)}{\tau_m} \\[6pt]
\nonumber \frac{d}{dt}\left[e^{t/\tau_m} v\right] &= \frac{e^{t/\tau_m} f(t)}{\tau_m} \\[6pt]
\nonumber e^{t/\tau_m} v + c_1 &= \frac{\displaystyle\int e^{t/\tau_m} f(t)\,dt}{\tau_m} = \frac{\displaystyle\int_0^t e^{t'/\tau_m} f(t')\,dt'}{\tau_m} + \frac{F(0)}{\tau_m} \\[6pt]
\nonumber e^{t/\tau_m} v &= c + \frac{\displaystyle\int_0^t e^{t'/\tau_m} f(t')\,dt'}{\tau_m} \\[6pt]
\nonumber v(t) &= c\,e^{-t/\tau_m} + \frac{e^{-t/\tau_m} \displaystyle\int_0^t e^{t'/\tau_m} f(t')\,dt'}{\tau_m} \\[6pt]
\nonumber v(t_0) = a &= c\,e^{-t_0/\tau_m} + e^{-t_0/\tau_m}\int_0^{t_0} e^{t'/\tau_m} f(t')\,dt' \\[3pt]
\nonumber \Rightarrow\quad c &= a\,e^{t_0/\tau_m} - \int_0^{t_0} e^{t'/\tau_m} f(t')\,dt' \\[6pt]
\nonumber v(t) &= \left(a\,e^{-t_0/\tau_m} - \int_0^{t_0} e^{t'/\tau_m} f(t')\,dt'\right) e^{-t/\tau_m} + e^{-t/\tau_m}\int_0^t e^{t'/\tau_m} f(t')\,dt' \\[6pt]
\nonumber v(t) &= a\,e^{-(t-t_0)/\tau_m} + \int_{t_0}^t e^{(t'-t)/\tau_m} f(t')\,dt' \\[6pt]
v(t) &= a\,e^{-(t-t_0)/\tau_m} + \int_0^{t-t_0} e^{-t'/\tau_m} f(t-t')\,dt'. 
\end{align}
Thus, given the above, we finish the construction of the LIF SRM as follows: 
\begin{align}
\nonumber \tau_m \frac{dv_i(t)}{dt} &= -v_i(t) + R\sum_j w_{i,j} \sum_k \alpha(t - t_{j,k}) + R\,I^{ext}(t) \\[10pt]
\nonumber &\text{with } a = v_i(t_{i,\hat{k}}) = v_{reset},\quad f = R\sum_j w_{i,j} \sum_k \alpha(t - t_{j,k}) + R\,I^{ext}(t) \notag\\[10pt]
\nonumber v_i(t) &= v_{reset}\,e^{-(t - t_{i,\hat{k}})/\tau_m} \notag\\
\nonumber &\quad + \sum_j w_{i,j} \sum_k \frac{1}{C}\int_0^{t - t_{i,\hat{k}}} e^{-t'/\tau_m}\,\alpha(t - t_{j,k} - t')\,dt' \notag\\
\nonumber &\quad + \frac{1}{C}\int_0^{t - t_{i,\hat{k}}} e^{-t'/\tau_m}\,I^{ext}(t - t')\,dt' \\[10pt]
\nonumber &= \eta(t - t_{i,\hat{k}}) + \sum_j w_{i,j} \sum_k \epsilon(t - t_{i,\hat{k}},\, t - t_{j,k}) \notag\\
&\quad + \int_0^{\infty} \kappa(t - t_{i,\hat{k}},\, t')\,I^{ext}(t - t')\,dt' . \label{eqn:derived_srm_appendix}
\end{align}
Finally, as remarked in the main paper, the SRM of Equation \ref{eqn:derived_srm_appendix} entails evaluating an integral, which itself will require the maintenance of a history/buffer of input event spike times.

\end{document}